%% file: main_for_review.tex
\documentclass[10pt,twocolumn,letterpaper]{article}

\usepackage{cvpr}
\usepackage{times}
\usepackage{epsfig}
\usepackage{graphicx}
\usepackage{amsmath,amssymb,subfigure}
\usepackage{algorithm,algpseudocode}
\usepackage{booktabs,multirow,comment}
\usepackage{breqn,caption,array,wrapfig}
\usepackage{balance,colortbl,enumitem}
\input{sections/defines}

\usepackage[font=small,skip=5pt]{caption}
\usepackage[pagebackref=true,breaklinks=true,letterpaper=true,colorlinks,bookmarks=false]{hyperref}

\cvprfinalcopy

\begin{document}

\title{
	A Novel Technique for Evidence based Conditional Inference \\ 
	in Deep Neural Networks via Latent Feature Perturbation  
}	

\author{Dinesh Khandelwal \quad Suyash Agrawal \quad Parag Singla \quad Chetan Arora \\
	Indian Institute of Technology Delhi 
}

\maketitle

\input{./sections/abstract.tex}
\vspace{-1em}
\input{./sections/intro.tex}
\input{./sections/related.tex}
\input{./sections/algo.tex}
\input{./sections/experiments.tex}
\input{./sections/conclusion.tex}
\nocite{shrivastava2016contextual}
{\small
\bibliographystyle{ieee_fullname}
\bibliography{all}
}

\appendix
\input{./sections/suppl.tex}

\end{document}

%% file: sections/defines.tex
\newcommand{\multilineC}[1]{\begin{tabular}[b]{@{}c@{}}#1\end{tabular}}

\def\xi{x^{(i)}}
\def\yi{y^{(i)}}
\def\ai{a^{(i)}}
\def\ah{\widehat{a}}
\def\yh{\widehat{y}}
\def\yhi{\yh^{(i)}}
\def\ahi{\ah^{(i)}}

\def\LS{\mathcal{L}}
\def\x{\mathbf{x}}
\def\Pr{\mathbb{P}}
\def\tmax{{\text{max}}}
\DeclareMathOperator*{\argmax}{arg\,max}

\algnewcommand\algorithmicinput{\textbf{Input:}}
\algnewcommand\INPUT{\item[\algorithmicinput]}
\algnewcommand\algorithmicoutput{\textbf{Output:}}
\algnewcommand\OUTPUT{\item[\algorithmicoutput]}
\renewcommand{\algorithmiccomment}[1]{\bgroup\hfill\small//~#1\egroup}
\algnewcommand{\LineComment}[1]{\State \bgroup\small//~#1\egroup}
\algnewcommand{\LeftComment}[1]{\Statex \bgroup\small//~#1\egroup}

\newcommand{\myparagraph}[1]{\vspace{0.5em} \noindent \textbf{#1}: }

%% file: sections/abstract.tex
\begin{abstract}
Auxiliary information can be exploited in machine learning models using the paradigm of evidence based conditional inference. 
%Presence of auxiliary informEvidence based conditional inference is a useful paradigm to perform inference in presence of auxiliary information.robust predictions in the presence of ambiguities. 
Multi-modal techniques in Deep Neural Networks (DNNs) can be seen as perturbing the latent feature representation for incorporating evidence from the auxiliary modality. However, they require training a specialized network which can map sparse evidence to a high dimensional latent space vector. Designing such a network, as well as collecting jointly labeled data for training is a non-trivial task. In this paper, we present a novel multi-task learning (MTL) based framework to perform evidence based conditional inference in DNNs which can overcome both these shortcomings. Our framework incorporates evidence as the output of secondary task(s), while modeling the original problem as the primary task of interest. During inference, we employ a novel Bayesian formulation to change the joint latent feature representation so as to maximize the probability of the observed evidence. Since our approach models evidence as prediction from a DNN, this can often be achieved using standard pre-trained backbones for popular tasks, eliminating the need for training altogether. Even when training is required, our MTL architecture ensures the same can be done without any need for jointly labeled data. Exploiting evidence using our framework, we show an improvement of $3.9$\% over the state-of-the-art, for predicting semantic segmentation given the image tags %level tags, 
and  $2.8$\% for predicting instance %level 
segmentation given image captions. %image captions. 

%Many prediction tasks, especially in computer vision, are often inherently ambiguous. While the ideal output of a semantic segmentation task depends on the scale one is looking at, image saliency is often user dependent. Similarly, the output of a video summarization or an image captioning task is often a function of the usage context. In classical machine learning, such situations are typically handled by providing evidence at the run time and then making predictions conditioned upon the evidence. Depending upon the case, the evidence may represent sample specific scale, saliency or context. With deep neural networks becoming the work horse for solving many computer vision problems, it is not clear how such evidences could be incorporated in the predictions made by these systems. In this paper, we propose a generic framework which allows one to use arbitrary types of per sample evidences at the test time, and then make predictions conditioned upon such evidences. We demonstrate the effectiveness of our framework by giving specific implementations for two scenarios: predicting semantic segmentation given the image type (image classification) as evidence, and predicting instance level segmentation given the text description of the image. 
\end{abstract} 

%% file: sections/intro.tex
\section{Introduction}

%Deep Neural Networks (DNNs) have become the work horse for solving many computer vision problems. Many times additional auxiliary information easily available at the prediction time for example, captions on a Facebook image and can be used both at training and test time to improve the performance of tasks like instance segmentation. In this paper, we address the question: How to use auxiliary information at prediction time to improve the performance of state-of-the-art DNNs for computer vision problems ? We propose a generic framework which allows one to use arbitrary types of per sample evidence at the test time, and then make predictions conditioned upon such evidence.

\begin{figure}[t]
	\begin{center}
		\includegraphics[width=\linewidth]{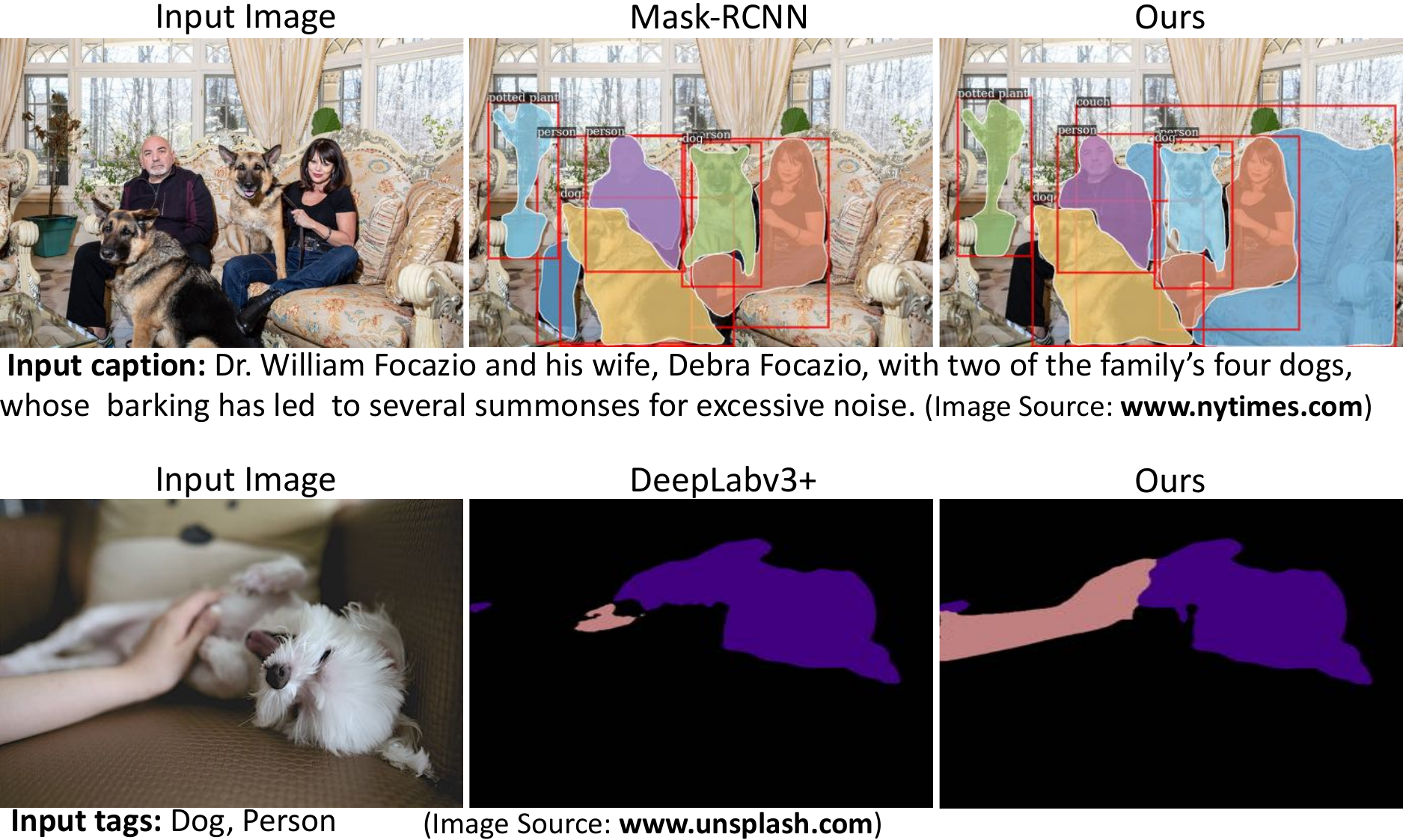}
		\caption{The figure shows some sample results by our framework exploiting the commonly available evidence at the test time. The 1st row shows the output of instance segmentation exploiting corresponding textual description, whereas the 2nd row gives semantic segmentation output using image level tags as the evidence (auxiliary information). Both the images as well as auxiliary information are deliberately sourced from unconstrained data on the web (and not the benchmark datasets) to show easy availability of such information at the test time.}
		\label{fig:real_world_examples}
	\end{center}	
\end{figure}

For many problems of interest, auxiliary information is often available at the inference time, e.g., presence of image tags while doing segmentation.
%or image captions while doing video summarization. 
Exploiting such information can presumably result in improved predictions.
%Many prediction tasks, especially in computer vision, are often inherently ambiguous. While the ideal output of a semantic segmentation task depends on the scale one is looking at, image saliency is often user dependent. Similarly, the output of a video summarization or an image captioning task is often a function of the usage context. 
In classical machine learning, such as probabilistic graphical models~\cite{koller2009probabilistic}, %these situations are 
such auxiliary information is typically handled by providing evidence at the run time and then making predictions conditioned upon the evidence. With DNNs becoming the work horse for solving many computer vision problems, it becomes important that such available evidence be incorporated in the predictions made by these systems too. %as well. 
In this paper, we propose a generic framework which allows one to use arbitrary types of per sample evidence at the test time, and then make predictions conditioned upon such evidence. Figure~\ref{fig:real_world_examples} shows some results from our framework.

One of the ways to incorporate evidence at test time in DNNs is by using multi-modal input~\cite{ngiam2011multimodal}, where the evidence is provided as additional input in the auxiliary modality. The techniques are also commonly referred to as `multi-stream' architectures~\cite{zolfaghari2017chained} or `fusion'~\cite{feichtenhofer2016convolutional} based approaches. The flip side of multi-modal architectures is that one needs huge amount of jointly labeled data to train the architecture. Another practical limitation is that many times the available evidence is sparse. For instance, when solving the problem of semantic segmentation with image tags as evidence, designing a multi-modal DNN which takes a $20$ dimensional image level tag vector as evidence, and merges it with a $32 \times 32 \times 256$ latent feature representation, from the primary task, becomes a non-trivial task. Another problem is the possibility of a DNN ignoring the low-dimensional latent code from the evidence in favor of the high-dimensional feature vector from the primary task. This has been reported in the case of conditional image-to-image GANs \cite{gan_mode_collapse}, where the generator tends to ignore the low-dimensional latent code in favor of the high-dimensional image, causing mode collapse.

One of the ways to look upon multi-modal techniques is the \emph{perturbation} induced in the latent feature representation. One could argue that there is an inherent latent feature representation learnt by the standard uni-modal architecture for the task. In the case of multi-modal architectures, this representation is \emph{adaptively perturbed} by using the auxiliary modality. The final inference is then made on the basis of this perturbed representation. The perturbation perspective motivates one to look at other ways of changing the latent representation in favor of evidence, which do not have the limitations of multi-modal DNNs. 

%Comparison with some earlier work ~\cite{rosenfeld2018priming} shows that our approach can be significantly more effective (see Section~\ref{sec:relatedwork},~\ref{sec:expt}).
%than multi-modal setting.
%a multi-modal based approach.
%(see Sections~\ref{sec:related},~\ref{sec:expt}).

The main contribution of this paper is in suggesting a Multi-Task Learning (MTL) \cite{caruana1995learning} based framework to perform evidence based conditional inference in DNNs. We model our task of interest as the primary task in the MTL framework. The evidence is modelled as the output of one or more auxiliary tasks. While our training process is identical to the one used by standard MTL architectures, our testing phase is quite different. We employ a novel Bayesian formulation to change the internal feature (joint) representation such that the probability of predicting the observed auxiliary task label(s) is maximized. We formulate this in terms of minimizing a loss over the auxiliary task label(s) and employ back propagation to arrive at the modified representation. Comparison with earlier work ~\cite{rosenfeld2018priming} shows that our approach can be significantly more effective than multi-modal setting.
%( Section~\ref{sec:relatedwork},~\ref{sec:expt}).
%than multi-modal setting.
Our contributions are as follows:
%Our framework has several advantages: 
\begin{enumerate}[leftmargin=*]

    \item We present a novel approach to model evidence as the prediction of a DNN in an MTL framework. Our approach allows us to make use of pre-trained DNN architectures for many standard tasks, and does not require jointly annotated data for training.

    %is easier to design compared to multi-modal setting, and allows us to make use of pre-trained DNN architectures for many standard tasks. Our approach does not require any jointly annotated data for training.
    %unlike existing approaches. 
    %due to the MTL setup. 
    %To the best of our knowledge, this is the first attempt to incorporate evidence at inference time in the given manner.
    \item We present a novel Bayesian formulation which maximizes the probability of evidence via latent feature perturbation. This provides a strong mathematical basis for our approach.
    %maximizes the probability of evidence in a Bayesian framework giving our approach a strong mathematical foundation. 

    \item We present experiments on three different tasks demonstrating that exploiting evidence using our method can result in significantly improved predictions over the state-of-the-art, as well as multi-modal baselines.
    %significant performance gains over state-of-the-art on 3 different tasks: (a) semantic segmentation (b) instance segmentation (c) image captioning, using image tags, image captions and image tags as the auxiliary information respectively.
\end{enumerate}

%% file: sections/related.tex
\section{Related Work} \label{sec:relatedwork}
One way to handle auxiliary information is to pose the problem in a multi-modal setting where the inputs are represented using multiple modes, and one (or more) of the inputs can be treated as evidence\cite{ngiam2011multimodal}. %For most part, existing work has focused on a setting where all the modes come from a dense feature representation. 
For example, \cite{simonyan2014two} has used optical flow between the consecutive frames as temporal evidence to improve the performance of action recognition in videos. \cite{BMVC2016_73} has improved pedestrian detection in RGB images using thermal images as evidence. 

%Very close to our work are 
Priming Neural Networks~\cite{rosenfeld2018priming} deal with a multi-modal setting while their approach allows sparse evidence, but it is restricted to CNNs only. One hot encoding of evidence is first passed through a linear layer which is then used to compute a modulated (weighted) additive combination of the features at multiple layers in the network. The weights of the linear transformation are learned while keeping the weights of the rest of the network fixed (which are learned apriori). On the task of semantic segmentation with object tags as evidence and on a slightly older version of dataset used in our experiments, they obtain a gain of less than 1\% compared to our gain of 3.9\% (see Section~\ref{sec:expt} for more details). We also implement a multi-modal technique more sophisticated than theirs, and demonstrate this stronger baseline also performs worse than our approach. \cite{kilickaya2016leveraging} focus on the task of improving object detection using caption information. Their idea can be effectively seen as a simple pruning based strategy, which did not work well for more sophisticated tasks of semantic/instance segmentation in our experiments.~\cite{cheng2014imagespirit} present an approach to incorporate human interactive feedback to improve the image segmentation results. Their focus is on involving a human in the loop which may not always be possible, as opposed to working with inert evidence information at inference time.

Posterior regularization~\cite{ganchev2010posterior} imposes a set of (soft) constraints on the output distribution, albeit, based on characteristics of the underlying data statistics. But no sample specific evidence is used. Some other recent works~\cite{pathak2015constrained, xu2017semantic, marquez2017imposing} has proposed learning a deep neural network in presence of constraints specified in the form of rules expressing domain knowledge. Lee et al. \cite{lee2017enforcing} has proposed to enforce test time constraints on a DNN output. While in most of these works, the idea is to enforce `prior deterministic constraints' arising out of natural rule based processing, our framework is inspired from using any easily available and arbitrary type of auxiliary information. 

Our work is also different from multi-task learning~\cite{liang2015semantic, bingel2017identifying, romera2012exploiting}, and weakly supervised learning~\cite{pathak2015constrained,khoreva2017simple} paradigms, which make use of auxiliary information only during training time. In weakly supervised learning, only partial supervision is provided in the form of auxiliary task labels during training. In contrast, we make full use of training labels, and also exploit the auxiliary information during inference.

%% file: sections/algo.tex
\begin{figure}[t]
	\centering
	\subfigure{\includegraphics[width=\linewidth]{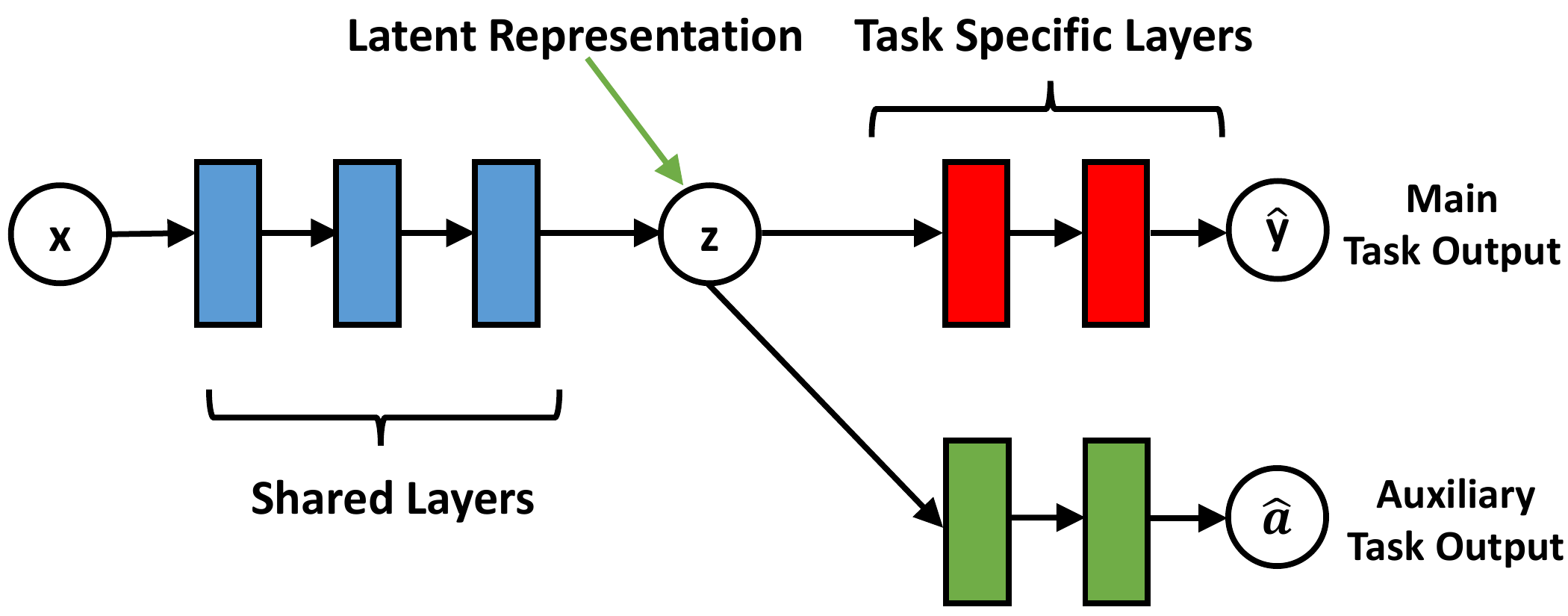}}
	\caption{Model Architecture for Multi-task learning (MTL)}
	\label{fig:model_architecture}
\end{figure}

\section{Proposed Approach} \label{sec:approach}

In the following section, we first describe our framework at the train time which is formulated as an MTL problem. We then describe theoretical foundation of the latent vector perturbation, followed by our actual implementation based on backpropagating loss from the auxiliary task at test time.

\subsection{Proposed Architecture}

Our solution employs a generic MTL~\cite{ruder2017overview} based architecture which consists of a main (primary) task of interest, and another auxiliary task whose desired output (label) represents the evidence in the network. 

\vspace{0.75em}

\noindent \textbf{Notation:} 
We use $P$ to denote the primary task of interest, and $A$ to denote the auxiliary task in the network. Let $(\xi,\yi,\ai)$ denote the $i^{\text{th}}$ training example, where $\xi$ is the input feature vector, $\yi$ is the desired output (label) of the primary task, and $\ai$ denotes the desired output (label) of the auxiliary task. Correspondingly, let $\yhi$ and $\ahi$ denote the output produced by the network for the primary task and auxiliary tasks, respectively. 

\vspace{0.75em}

\noindent \textbf{Model:} 
Figure~\ref{fig:model_architecture} shows the MTL based architecture~\cite{ruder2017overview} for our set-up. There is a common set of layers shared between the two tasks, followed by the task specific layers. $z$ represents the common hidden feature representation fed to the two task specific parts of the architecture. For ease of notation, we will refer to the shared set of layers as {\em trunk}. The network has three sets of weights. First, there are weights associated with the trunk denoted by $W_{zx}$. $W_{yz}$ and $W_{az}$ are the sets of weights associated with the two task specific branches, primary and auxiliary, respectively. The total loss $\LS_{T}(\cdot)$ is a function of these weight parameters and is defined as:
\[
\LS_{T}(\cdot)= \sum\nolimits_{i=1}^{m} \LS_{P} \Big( \yhi,\yi \Big)
			  + \lambda \sum\nolimits_{i=1}^{m} \LS_{A} \Big( \ahi,\ai \Big) \nonumber
\]
Here, $\LS_{P}(\cdot)$ and $\LS_{A}(\cdot)$ denote the loss for the primary and auxiliary tasks, respectively, and $\lambda$ is the importance weight for the auxiliary task. The sum is taken over $m$ examples in the training set. $\LS_{P}$ is a function of the shared set of weights $W_{zx}$, and the task specific weights $W_{yz}$. Similarly, $\LS_{A}$ is a function of the shared weights $W_{zx}$ and task specific weights $W_{az}$.

\vspace{0.75em}

\noindent \textbf{Training:} 
The goal of training is to find the weights which minimize the total loss over the training data. Using the standard approach of gradient descent, the gradients can be computed as follows:
\begin{align}
\nabla_{W_{yz}} \LS_{T}(\cdot) & = \sum\nolimits_{i=1}^{m} \nabla_{W_{yz}} \LS_P(\cdot), \\
\nabla_{W_{az}} \LS_{T}(\cdot) & = \sum\nolimits_{i=1}^{m} \nabla_{W_{az}} \LS_A(\cdot), \text{ and} \\
\nabla_{W_{zx}} \LS_{T}(\cdot) & = \sum\nolimits_{i=1}^{m} \nabla_z \LS_P(\cdot) \nabla_{W_{zx}}(z) \nonumber \\
						   & + \lambda \sum\nolimits_{i=1}^{m} \nabla_z \LS_A(\cdot) \nabla_{W_{zx}} (z) \label{eq:joint_training}
\end{align}
Note that the weights in the task specific branches, i.e., $W_{yz}$ and $W_{az}$, can only affect losses defined over the respective tasks (Eqs. 1 and 2 above). On the other hand, weights $W_{zx}$ in the trunk affect the losses defined over both the primary as well as the auxiliary tasks. 

\subsection{A Bayesian Framework to Incorporate Evidence in DNNs}

\setlength{\columnsep}{5pt}%
\setlength{\intextsep}{5pt}%
\begin{wrapfigure}{r}{0.4\linewidth}
\centering
\includegraphics[width=\linewidth]{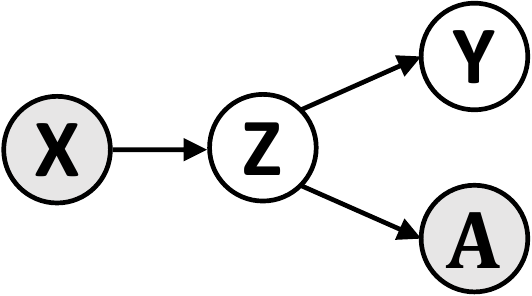}
\end{wrapfigure}
Consider the Bayesian network on the right, modeling the dependency between $X$ (input), $Y$ (primary output) and $A$ (auxiliary output). Here, $X$ and $A$ are observed variables at the test time and are shown in gray. The Bayesian network (BN) abstracts out the dependencies in the MTL architecture given in Figure~\ref{fig:model_architecture}. During inference time, we are interested in finding the best value of $y$ conditioned on $x$, given the DNN weights which essentially parameterize the probability function. In the standard setting, both $a$ and $y$ are unknown at the test time, and therefore, we are interested in finding the $y$ such that: $\Pr(y|x)$ is maximized. This can be written as:
\[ 
\begin{aligned}
    \argmax\nolimits_y \Pr(y|x) & = \argmax\nolimits_y \sum\nolimits_z \Pr(y|z) \Pr(z|x) \\
    				   & \approx \argmax\nolimits_y \Pr(y|z_\tmax) \Pr(z_\tmax|x) \\
    				   & \approx \argmax\nolimits_y \Pr(y|z_\tmax)
\end{aligned} 
\]
Here, $z_\tmax$ denotes the value of $z$ maximizing $\Pr(z|x)$. It is important to note that we have assumed a distribution over variable $z: \Pr(z|x)$ (e.g., Gaussian) with the DNN outputting the mode of the distribution. Therefore, $z_\tmax$ as well as $\Pr(y|z_\tmax)$ can be simply obtained by forward propagation using the DNN weights, as done in the standard inference. In our setting, we have additional evidence from the auxiliary task available to us. Therefore, the problem of finding the optimal $y$ now changes to finding the $y$ such that: $P(y|x,a)$ is maximized:
\begin{align}
    \argmax\nolimits_y \Pr(y|x,a) & = \argmax\nolimits_y \sum\nolimits_z \Pr(y|z) \Pr(z|x,a) \nonumber \\ 
    & \approx \argmax\nolimits_y \Pr(y|z'_\tmax) \Pr(z'_\tmax|x,a) \nonumber \\
    & \approx \argmax\nolimits_y \Pr(y|z'_\tmax) \label{eq:graphical_model}
\end{align}
Here, $z'_\tmax$ denotes the value of $z$ maximizing $\Pr(z|x,a)$. Note that $Y$ is independent of $A$ given $Z$ in our BN, therefore: $\Pr(y|z,a)=\Pr(y|z)$. 

$P(y|z'_\tmax)$ in Eq. (\ref{eq:graphical_model}) can be computed using forward propagation as before over the DNN parameters $W_{yz}$, assuming we know the value of $z'_\tmax$. The issue lies in computing the value of $z'_\tmax$ since it is now conditioned on both $x$, and $a$. Therefore, we are interested in finding: $\argmax_z \Pr(z|x,a)$. In other words, what is the optimum value of $z$ (feature representation) given the observed information about the auxiliary task in addition to input $x$. We propose two approaches:
\begin{enumerate}[leftmargin=*]
	\item \textbf {Z-Direct (ZD):} In the first proposal, we compute $\partial \LS_A(\ah,e) / \partial z$. Here $\ah$ is the output of the DNN for the auxiliary task, whereas $e$ is the observed evidence. We perturb the value of $z$ opposite to the computed gradient direction with the learning rate $\alpha$. After the update, the new value of $z$ is guaranteed to produce the $\ah$ which is closer to the observed evidence $e$. The process is repeated until we find the $z'$ which produces the observed evidence. 
	\item \textbf {Z-Weight (ZW):} The $z$ obtained from the procedure outlined above may be infeasible in the sense that there is no valid mapping from $x$ to $z$. Therefore, in the second proposal we instead suggest to change the weight of the trunk, $W_{zx}$, as per the gradient direction obtained from $\partial \LS_A(\ah, e) / \partial W_{zx}$. Similar to the above we repeat the update process until we arrive at the value of $z'$ which produced the observed evidence $e$.
\end{enumerate}
In theory the methods described above can be used as-is to find $\argmax_z \Pr(z|x,a)$. However, in practice, we need a regularizer during the update process to avoid over-fitting $z$ to the observed evidence. This is described in the next section.
The idea to perturb latent feature representation described above, bears similarity to the Cross-Grad~\cite{shankar2018generalizing}, where they perturb the input signal to include unrepresented domains in the training data. Whereas in their case, perturbation is only done during training time, we take this idea further to perturbation of our feature representation $z$ at prediction time to suitably match the observed evidence.

It may be noted that, though our formulation has been described in terms of a single auxiliary task, it is straightforward to extend it to a setting with more than one auxiliary task (and associated evidence at the inference). We show experiment with multiple auxiliary tasks in Section~\ref{sec:multiple_evidence_simultaneously}.

\begin{figure}
\centering
\subfigure{\includegraphics[width=\linewidth]{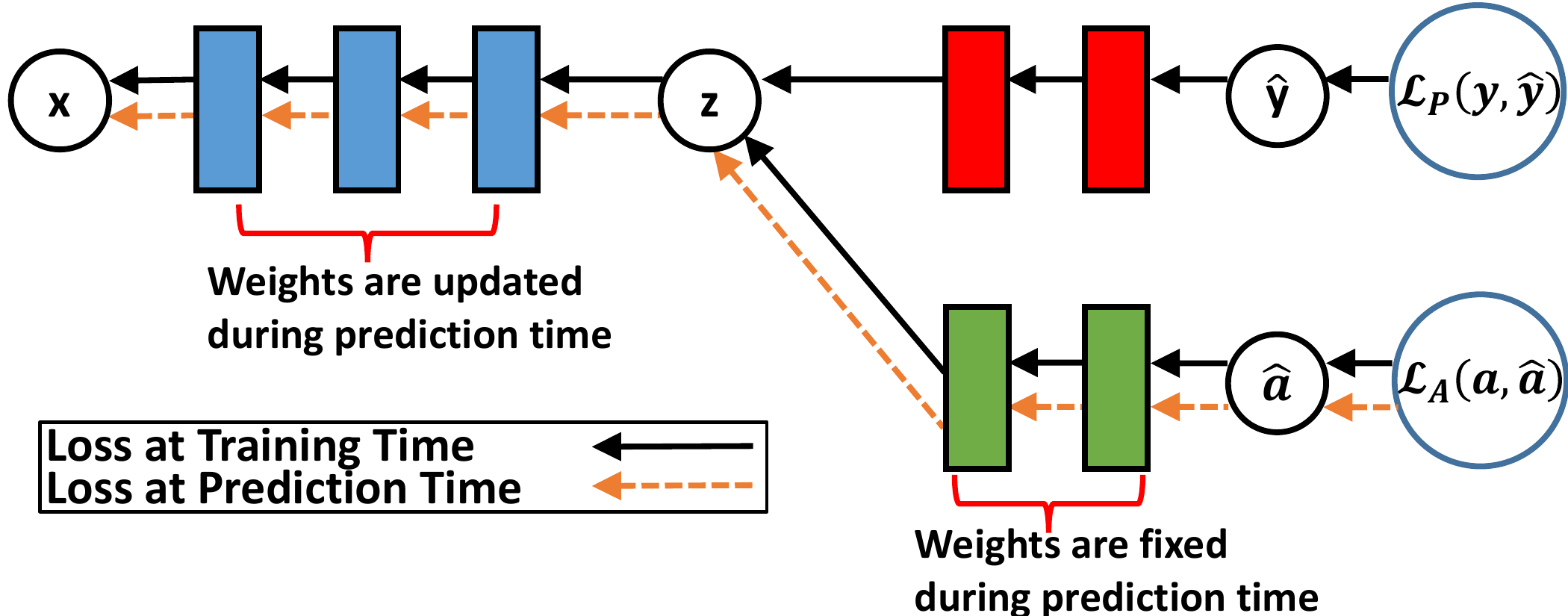}}
\caption{Loss propagation at train and prediction time}
\label{fig:loss_propagation}			
\end{figure}

\subsection{Optimization using Back-Propagation}
We present our approach for back-propagating the loss over auxiliary task labels during prediction time. During test time, we are given additional information about the output of the auxiliary task. Let us denote this by $e$ (evidence) to distinguish it from the auxiliary outputs during training time. We define a loss in terms of $\LS_A(\hat{a},e)$ and then back-propagate its gradient through the network. We also add a regularization term so that the new weights do not deviate much from the originally trained weights (to avoid over-fitting on $e$). Since the network weights $W_{az}$ in the auxiliary branch are fixed, the change in the proposed loss can be affected only through a change in the weights $W_{zx}$ in the trunk. In particular, the weights $W_{yx}$ and $W_{za}$ remain unchanged during this process. The gradient of the total loss can be re-written as:
%\vspace{-0.03in}
\[
\nabla_{W_{zx}} \LS_{\text{Test}}(\cdot) = \nabla_z \LS_A(\ah,e) \nabla_{W_{zx}}(z) + \beta \lVert W_{zx}-W^*_{zx} \rVert 
\]
%\vspace{-0.03in}
Here, %$W_{az}^{*}$ and 
$W_{zx}^*$ denote the weights learned during training and $\beta$ is the regularization parameter. Note that these equations are similar to those used during training (Eq. \ref{eq:joint_training}), with the differences that (1) The loss is now computed with respect to the single test example (2) Effect of the term dependent on primary loss has been zeroed out. (3) A regularizer term has been added. In our experiments, we also experimented with \emph{early stopping}, after a fixed number of updates, instead of adding the norm based regularizer.

Figure \ref{fig:loss_propagation} explains the weight update process pictorially. Once the new weights are obtained, they can be used in the forward propagation to obtain the desired value $\yh$ on the primary task. Algorithm~\ref{algo:weight_update} below describes the overall algorithm for weight update during test time.
\begin{algorithm}[h]
	%\algsetup{linenosize=\small}
	%\scriptsize
	\begin{algorithmic}[1]
		\INPUT $\x$ (input), $e$: evidence 
		\INPUT $\eta$ (Learning rate), $T$ (Iterations),
		\INPUT $W^{*}_{zx}$: Originally trained weights 
		\OUTPUT $W_{zx}$ 
		\For {$t \in \{1,\dots,T\}$};
		\State Calculate the loss $\LS_{\text{test}}(\ah,e)$ over evidence
		\State Compute $\nabla_{W_{zx}} L_{\text{test}}(\cdot)$, using back-propagation
		\State Update $W_{zx}$ using gradient descent rule
		\EndFor
		\State Return the newly optimized weights		
	\end{algorithmic}
	\caption{Weight Update Algorithm}
	\label{algo:weight_update}
\end{algorithm}

The proposal where the latent feature representation $z$ is changed directly, instead of weights $W_{zx}$ can be implemented in the same way.

%% file: sections/experiments.tex
\section{Experiments}\label{sec:expt}

We validate the effectiveness of our framework in two settings: semantic segmentation using image level tags, and instance segmentation using image captions. To show that our framework can work with multiple auxiliary information, we have also experimented with instance segmentation given both image level tags, as well as image captions. To demonstrate the capability of our framework beyond the segmentation task, we have also experimented with the task of image captioning given the image level tags. For each of the experimental settings, we have used popular state-of-the-art backbones for the respective tasks, and show the benefits of exploiting evidence over the same state-of-the-art as the baseline. Designing a multi-modal DNN incorporating sparse evidence is a non-trivial task, and we are not aware of any reported results for these configurations. Therefore, for comparison, we have designed our own multi-modal DNNs for these tasks. In the following sections, we describe the results for each of the settings.

\subsection{Image Captioning given Image Level Tags}

We first describe our experiments with image caption prediction as the primary task and image tags as the auxiliary task. We use the experiments to also do ablation study of different latent feature update proposals, viz, direct update or by updating weights $W_{zx}$. See the previous section for the details on the two. We have conducted experiments on the MS COCO dataset. The base architecture was ``Show, Attend and Tell''(SAT)~\cite{xu2015show}, one of the prominent image captioning model.  To use image level tags, we branch off after the encoder module and design a multi-class classifier. The resultant feature map is passed through an average pooling layer, a fully connected layer, and then finally a sigmoid activation function to get probabilty for each of the 80 classes. \begin{figure}[t]
	\centering
	\subfigure{\includegraphics[width=\linewidth]{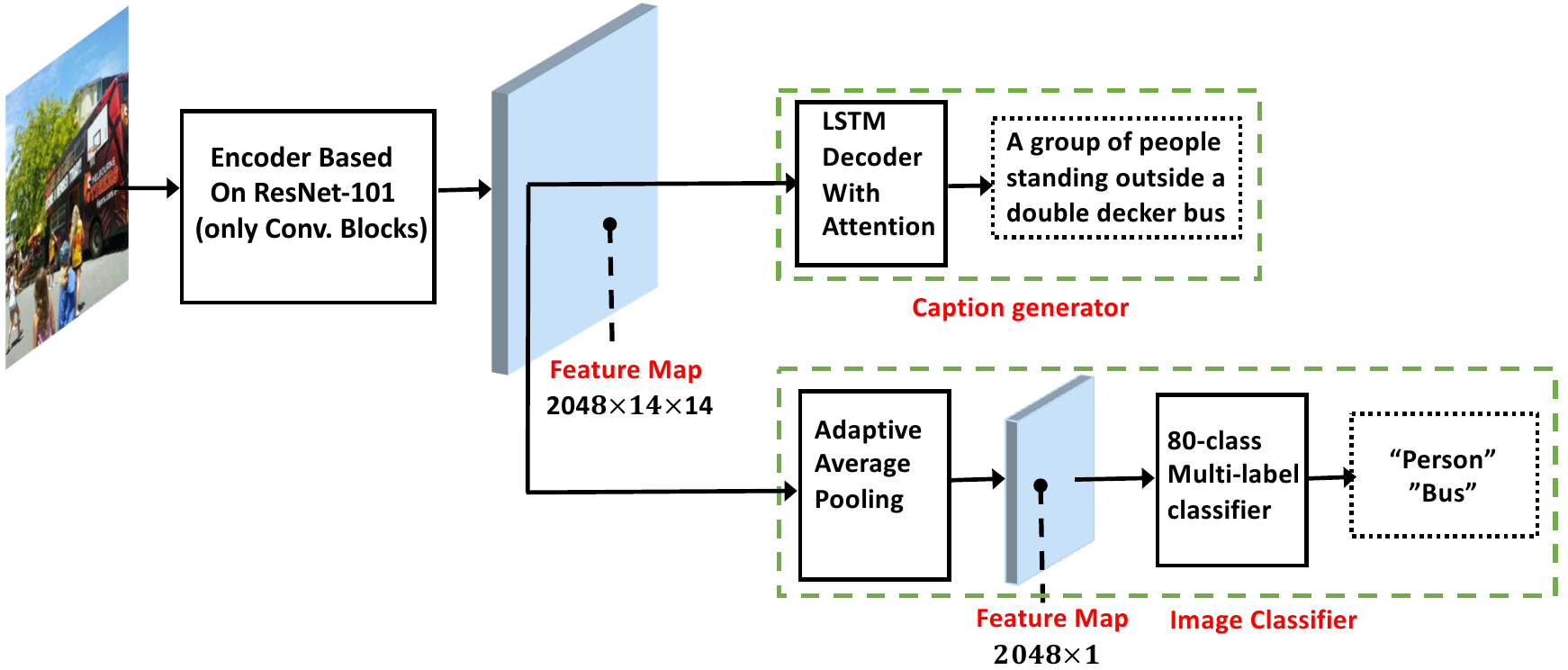}}
	\caption{Our framework for image captioning using ``Show, Attend and Tell''(SAT)~\cite{xu2015show} and a multiclass image classifier to use image level tags at the test time.}
	% Note that BB refers to bounding box.}
	\label{fig:caption-tags-proposed-arch}
\end{figure}
Figure~\ref{fig:caption-tags-proposed-arch} describes our proposed MTL architecture for the task. 

Using BLEU-4 as the metric, comparison results are shown in the table on the right below. Here, `S' refers to the baseline show-attend-and-tell approach, whereas the S-MT refers to the same trained in the MTL architecture with image classification as the auxiliary task. 
\setlength{\columnsep}{5pt}%
\setlength{\intextsep}{5pt}%
\setlength{\tabcolsep}{3pt}
\begin{wraptable}{r}{0.6\linewidth}
	\centering
	\begin{tabular}{lcccc}
		\toprule[1.5pt]
		& \textbf{S} & \textbf{S-MT} & \textbf{ZW} & \textbf{ZD}\\
		\midrule
		\textbf{BLEU-4} & 25.8 &  27.4  & 29.6 & 29.2 \\       
		\bottomrule[1.5pt]
	\end{tabular}
\end{wraptable}
Both `ZD', and `ZW' refer to our approaches, in which the latent vector is perturbed directly (ZD), or indirectly by changing the weights $W_{zw}$ (ZW). It is clear that, though both the proposed update mechanisms show improvement over state of the art, ZW performs slightly better than ZD. A similar behavior is observed in other experiments as well. We report all the following results using ZW update.
	
\input{./sections/results_semantic_seg.tex}

\input{./sections/results_instance_seg.tex}

\subsection{Using Multiple Evidence Simultaneously}\label{sec:multiple_evidence_simultaneously}

It is straightforward to use multiple types of auxiliary information simultaneously in our framework. We demonstrate the capability in the instance segmentation setup, where, in addition to using captions as the auxiliary information in the previous subsection, we now add image tags also as the second source of auxiliary information. Our proposed architecture for this task is a combination of the ones described in previous two subsections for semantic and instance segmentation. For exact details of the architecture please see the supplementary material. The model was trained in a MTL setting with one primary and two auxiliary tasks. The evidence over the two auxiliary tasks was propagated simultaneously during test time. Table~\ref{table:instance_segmentation_caption_tags_comparison} shows the comparison using AP (Average Precision) as the metric.

%Figure~\ref{fig:instance-seg-captions-tags-proposed-arch} describes our MTL architecture which i

\setlength{\columnsep}{5pt}%
\setlength{\intextsep}{5pt}%
\setlength{\tabcolsep}{5pt}
\begin{table}
\centering
\begin{tabular}{lcccc}
	\toprule[1.5pt]
% 	& \multirow{2}{*}{\multilineC{\bf MRCNN}} &
% 	\multicolumn{2}{c}{\bf Ours} \\  
% 	\cmidrule{3-4} 
% 	& & {Ours-C} & {Ours-C+T} \\ 

 \textbf{} & \textbf{MRCNN} & \textbf{Ours-Cap} & \textbf{Ours-Cap-Tag}\\	\midrule
	AP & 39.8 &  40.5  & 40.9  \\       
	\bottomrule[1.5pt]
\end{tabular}
\caption{Comparison of results for instance segmentation with multiple types of auxiliary information. MRCNN represent baseline pretrained network Mask R-CNN. 
Ours-Cap and Ours-Cap-Tag represent Mask-RCNN with captions only and Mask R-CNN with captions and Tags both respectively.}
\label{table:instance_segmentation_caption_tags_comparison}
\end{table}

% \begin{figure}[t]
% 	\centering
% 	\subfigure{\includegraphics[width=\linewidth, height=0.7\linewidth]{./figures/instance-seg-captions-tags-proposed-arch.pdf}}
% 	\caption{Our framework, for instance segmentation with multiple types of auxiliary information simultaneously. }
% 	% Note that BB refers to bounding box.}
% 	\label{fig:instance-seg-captions-tags-proposed-arch}
% \end{figure}

\subsection{Failure analysis} 
\begin{figure}[t]
	\begin{center}
		\includegraphics[width=\linewidth]{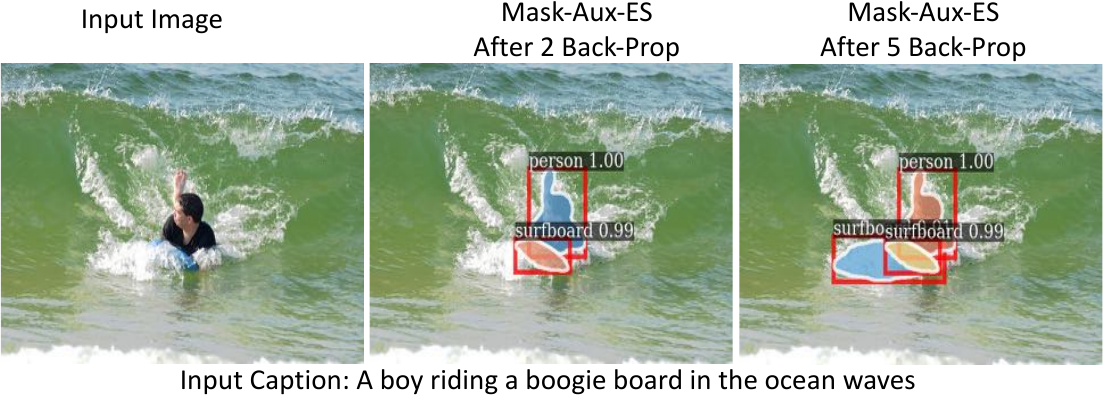}
		\caption{A failure case of our framework due to over-fitting on the auxiliary information when the number of back-propagation iterations are increased from 2 to 5.}
		\label{fig:error_analysis_sem_seg}
	\end{center}	
	%\hfill
\end{figure}

Overuse of back-propagation at test time may lead to over-fitting on the given auxiliary information in the proposed framework. Figure~\ref{fig:error_analysis_sem_seg} shows a failure example for our approach. As the number of back-propagation increases from 2 to 5, we observe over-fitting, leading to the prediction of multiple surf-boards. 

%% file: sections/results_semantic_seg.tex
\begin{figure}[t]
	\centering
	\includegraphics[width=\linewidth]{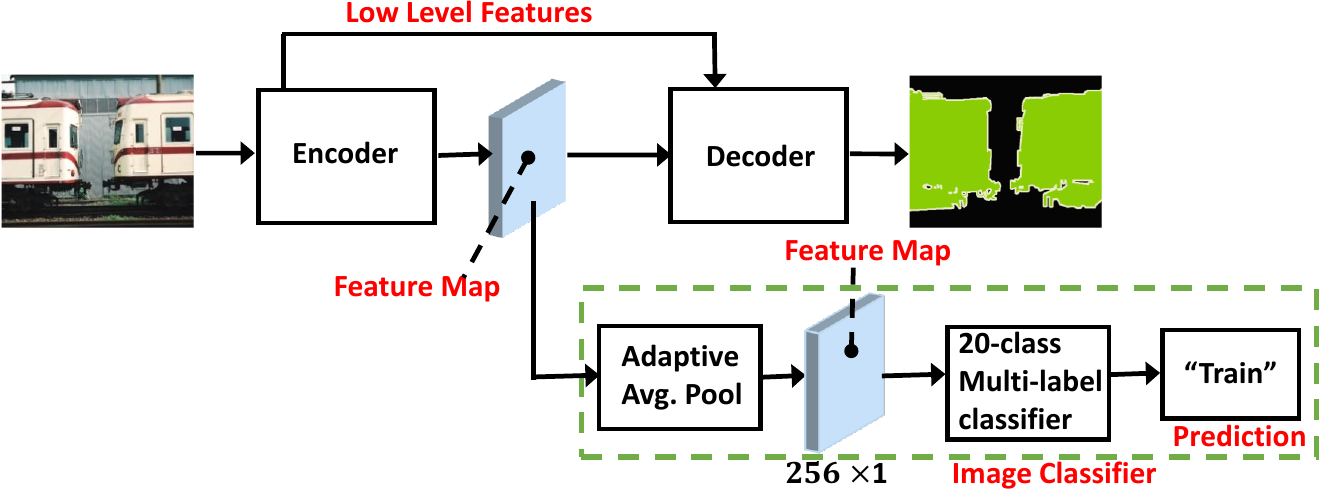}
	\caption{Our architecture for semantic segmentation using image level tags as test time evidence.}
	\label{fig:proposed-sem-seg-arch}
\end{figure}

\subsection{Semantic Segmentation}
\label{sec:semantic_seg}

The task of semantic segmentation involves assigning a label to each pixel in the image from a fixed set of object categories. Semantic segmentation is an important part of scene understanding and is critical first step in many computer vision tasks. In many semantic segmentation applications, image level tags are often easily available and encapsulate important information about the context, scale and saliency. We explore the use of such tags as auxiliary information at test time for improving the prediction accuracy. As clarified in earlier sections as well, though using auxiliary information in the form of natural language sentences \cite{hu2016segmentation, liu2017recurrent} have been suggested, these earlier works have used this information only during the training time. This is unlike us where we are interested in exploiting this information both during training as well as test.

\myparagraph{Evaluation Dataset}
For evaluation, we use PASCAL VOC 2012 segmentation benchmark \cite{everingham2010pascal}. It consists of 20 foreground object classes and one background class. We further augmented the training data with additional segmentation annotations provided by Hariharan et al.~\cite{hariharan2011semantic}. The dataset contains 10582 training and 1449 validation images. We use mean intersection over union (mIOU) as our evaluation metric which is a standard for segmentation tasks. 

\myparagraph{Our Implementation}
Our implementation builds on the publicly available official implementation of DeepLabv3+ \cite{chen2018encoder} available at \cite{deeplab-code}. For ease of notation, we will refer the DeepLabv3+ model as `DeepLab'. To use our framework, we have extended the DeepLab architecture to simultaneously solve the classification task in an MTL setting. Figure \ref{fig:proposed-sem-seg-arch} describes our MTL architecture. Starting with the original DeepLab architecture (top part in the figure), we branch off after the encoder module to solve the classification task. The resultant feature map is passed through an average pooling layer, a fully connected layer, and then finally a sigmoid activation function to get probabilty for each of the 20 classes (background class is excluded). 

For training, we make use of cross-entropy based loss, for the primary and binary cross entropy loss for the secondary task. %We first train the segmentation only network to get the initial set of weights.
We use pre-trained Deeplab model~\cite{deeplab-weights} to initialize the weights in the MTL based architecture (for the segmentation branch). The weights in the classification branch are randomly initialized. This is followed by a joint training of the MTL architecture. During prediction time, for each image, we back-propagate the binary cross entropy loss based on observed evidence over the auxiliary task (for test image) resulting in weights re-adjusted to fit the evidence. These weights are used to make the final prediction (per-image). The parameters in our experiments were set as follows. During training, the $\lambda$ parameter controlling the relative weights of the two losses is set of $1$ in all our experiments. During prediction, number of early stopping iterations was set to $2$. $\alpha$ parameter for weighing the two norm regularizer was set to $1$. We have used SGD optimizer with learning rate of $7* 10^{-4}$ at test time.

\setlength{\tabcolsep}{20pt}
\begin{table}
	\centering
	\begin{tabular}{lcc}
		\toprule[1.5pt]
		\textbf{Method} & \multicolumn{2}{c}{\bf mIoU} \\  \midrule & { No-MS} & {MS} \\ 
		\toprule[1.5pt]
		DL       & 82.45 & 83.58  \\
		DL-MT    & 82.59 & 83.55  \\
		DL-Prn    & 84.47 & 85.29  \\
		MM       & 83.66 & 85.12  \\
		\midrule
		Ours-L2 & 85.97 & 86.88  \\
		Ours-ES & 86.01 & 86.84  \\
		%		DL-BP-L2-Pr & 86.10& 87.00     \\
		%		DL-BP-ES-Pr & \textbf{86.22} & \textbf{87.03}    \\
		\bottomrule[1.5pt]
	\end{tabular}
\caption{Comparison of results for semantic segmentation. Two columns shows results of various approaches with and without using multi scale (MS) strategy. Please see the paper text for details on the configurations.}
\label{table:segementation_comparison}
\end{table}

\myparagraph{Compared Models} Table~\ref{table:segementation_comparison} compares the performance of our approach with various approaches. The model `DL' uses vanilla DeepLab based architecture, whereas `DL-MT' is trained using the MTL framework but do not use the evidence at the test time. One of our naive alternative is to simply prune or zero the probability of the label at each pixel which is not present in the tag set. The suffix `Prn' to refers to such alternative. We have also designed our own multi-modal network, referred to as MM, 
%which takes RGB image and image lebels as input.
%Our designed for MM 
taking inspiration from \cite{rosenfeld2018priming}.
% but somewhat more sophisticated than it. 
We take image features of size $a \times b \times c$ at an intermediate layer as shown in Figure~\ref{fig:proposed-sem-seg-arch} and then broadcast the
one-hot encoding of image labels which is of size 20 to every position in height and width. The resulting features %dimensions become $a\times b \times (c+20)$, which are than
are then passed through a $1\times1$ convolution to bring the final dimension back to $a \times b \times c$. The rest of the network is the same as DeepLabv3+. \cite{rosenfeld2018priming} has reported less than $1\%$ of improvement on an older version this dataset compared to our $3.9\%$ on the same task.

We experiment with two variations of our approach based on the choice of regularizer during prediction: ES uses early stopping and L2 uses an L2-norm based penalty. Original DeepLab paper~\cite{chen2018encoder}  also experiments with inputting images at multiple scales i.e, \{$0.5$,$0.75$,$1.0$,$1.25$,$1.5$\} as well as using left-right flipped input images, and then taking a mean of the predictions. We refer to this as multi-scaling (MS) based approach. We compare each of our models with multi-scaling being on and off in our experiments.

\input{./sections/table-sem-seg-classwise-miou.tex}

\myparagraph{Quantitative Results}
Table~\ref{table:seg_obj_comparision} presents the results for each of the object categories. For all the object categories except two, we perform better than the baselines. The gain is as high as 10 points (or more) for some of the classes. 

\myparagraph{Impact of Noise} 
Table~\ref{table:sem_noise_analysis} examines the performance of various approaches in the presence of noisy evidence. 
%\begin{comment}
%\setlength{\columnsep}{5pt}%
%\setlength{\intextsep}{5pt}%
\setlength{\tabcolsep}{10pt}
\begin{table}
	\centering
	\begin{tabular}{lcccc}
		\toprule[1.5pt]
		\textbf{Model} & \multicolumn{4}{c}{\textbf{\# of noisy tags ($k$)}} \\
		\midrule
		& \textbf{0}   & \textbf{1}   & \textbf{2}   & \textbf{3}     \\
		\midrule
		Ours & 86.0 & 85.3  &  84.0 & 83.4         \\
		Pruning  & 84.5 & 83.9  & 82.5 & 82.2 \\
		MM       & 83.7 & 83.7  & 83.4  & 83.2 \\
		\bottomrule[1.5pt]   
	\end{tabular}
\caption{Performance of different models as noisy tags are added in the auxiliary information. Multi-scaling was off in this expt.}
%in this experiment.}
\label{table:sem_noise_analysis}
\end{table}
%\end{comment}
For each image we randomly introduce $k$ (for varying values of $k$) additional (noisy) tags which were not part of the original set\footnote{This experiment was done without using multi-scaling}. Our approach is fairly robust to such noise, and its performance degrades slowly with increasing amount of noise. Even when we have 3 additional noisy tags added, we still do better than the baseline DeepLab models (Table~\ref{table:segementation_comparison}).
%DeepLab model.
%with a performance of 82.59.

\begin{figure}[t]
	\centering
	\subfigure{\includegraphics[width=\linewidth]{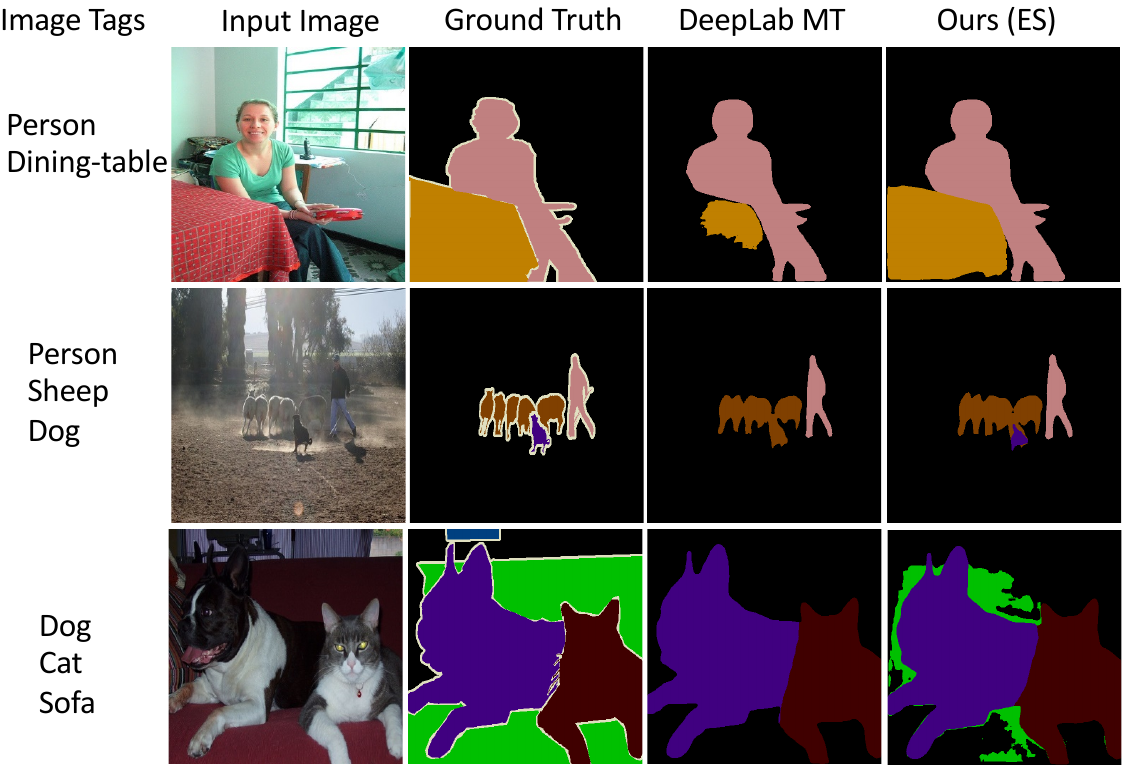}}
	\caption{Comparison of visual results on semantic segmentation using image level tags as the test time auxiliary information.}
	\label{fig:sem-seg-vis-output-main}
\end{figure}

\myparagraph{Visual Results}
Figure \ref{fig:sem-seg-vis-output-main} show the visual comparison of results for a set of hand picked examples. We note that our model is not only able to enhance the segmentation quality of already discovered objects, it is also able to discover new objects which are completely missed by the baseline. We show more results in the supplementary material.

\myparagraph{Inference Time}
Inference time on one image for different models are: $0.15$ seconds for DeepLab, $2.6$ seconds for Deeplab with multi-scaling, and $1.4$ seconds for our model with early stopping of 2 back-propagation. Our model with multi-scaling takes $3.85$ seconds on a GTX 1080 Ti GPU. 

\begin{comment}
\begin{figure}[t]
	\centering
	\subfigure{\includegraphics[height=0.35\linewidth]{./figures/mIoU_vs_backprop.pdf}} \;
	\subfigure{\includegraphics[height=0.35\linewidth]{./figures/mIoU_vs_reg.pdf}}		
	\caption{Sensitivity of results on number of back-propagation iterations (early stopping) and $\alpha$ parameter (L2-norm)}
	\label{fig:sem-seg-sensitivity}	
\end{figure}
\end{comment}

\myparagraph{Sensitivity Analysis}
%Figure \ref{fig:sem-seg-sensitivity} presents the sensitivity analysis with respect to the number of early stopping iterations (left) and the parameter controlling the weight of the L2 regularizer during prediction (right). There is a large range of values where we get significant improvements. 
%
We have also done sensitivity analysis with respect to the number of early stopping iterations and the parameter controlling the weight of the L2 regularizer during prediction. There is a large range of values where we get significant improvements. The results are given in the supplementary material.

%% file: sections/table-sem-seg-classwise-miou.tex
\setlength{\tabcolsep}{3pt}
\begin{table*}[t]
	\resizebox{0.78\textwidth}{!}{\begin{minipage}{\textwidth}
			\begin{tabular}{p{3cm}lllllllllllllllllllll}
				\toprule				
				
				\textbf{Model} & \textbf{bike} &	\textbf{table} &	\textbf{chair} &	\textbf{sofa} & \textbf{plant} &	\textbf{boat}	& \textbf{tv} &	\textbf{bottle} &	\textbf{bird} &	\textbf{person} &	\textbf{mbike} &	\textbf{car} &	\textbf{aero} &	\textbf{dog} &	\textbf{horse}	& \textbf{sheep} &	\textbf{train} &	\textbf{cat} &	\textbf{bg} &	\textbf{cow} &	\textbf{bus} \\ 
				
				\midrule
				
				\textbf{Pruning} & 45.4	& 62.1 &	62.5&	66.2 &	68.5 &	81.1 &	82.1 &	83.5 &	89.7 &	90.7 &	90.8 &	92.2 &	93.1 &	95.3 &	94.5 &	95.5 &	95.5 &	96.2 &	95.8 &	96.5 &	96.9 \\
				
				\textbf{MM} & 44.2 &	68.3 &	59.3 &	\textbf{79.8} &	65.7	& 70.9 &	80.9 &	83.3 &	89.7 &	90.3 &	91.6 &	89.6 &	91.5 &	92.6 &	92.0 &	91.9 &	94.1 &	95.0	& 95.9 &	93.9 &	96.3 \\

				\textbf{Ours-ES} &  \textbf{45.9} &	69.2 &	67.7 &	\textbf{79.8}	& \textbf{75.2} & \textbf{84.1} &  \textbf{82.9} & 86.1 &	\textbf{90.3} &	\textbf{91.7} &	92.6 &	\textbf{93.7} &	\textbf{94.6} & \textbf{96.2}&	95.1 &	\textbf{96.1} &	95.1 &	\textbf{96.4} &	\textbf{96.5} &	\textbf{97.4} &	\textbf{97.1} \\

				\textbf{Ours-L2} & \textbf{45.9}	 & \textbf{69.9}	 & \textbf{68.2}	&  79.7	& 74.7 &	\textbf{84.1} &	82.8 &	\textbf{86.3} &	90.1 &	91.6 &	\textbf{92.7} &	\textbf{93.7} &	\textbf{94.6} & \textbf{96.2} &	\textbf{95.1} &	\textbf{96.1} &	95.2 &	\textbf{96.4}	& \textbf{96.5} &	\textbf{97.4} &	\textbf{97.1} \\			
								
				\bottomrule
			\end{tabular}
			
	\end{minipage}}
	\caption{Object category-wise comparison of results for semantic segmentation on Pascal VOC. Numbers denote mIoU.} 
		\label{table:seg_obj_comparision} 
\end{table*}

%% file: sections/results_instance_seg.tex
\subsection{Instance Segmentation}
\label{sec:instance_seg}

Next, we present our experimental evaluation on a multi-modal task of object instance segmentation given textual description of the image. In an instance segmentation problem the goal is to detect and localize individual objects in the image along with segmentation mask around the objects. In our framework, we model instance segmentation as the primary task and image captioning as the auxiliary. Arguably, instance segmentation is more challenging than semantic segmentation, and incorporating caption information also seems significantly harder due to its frequent noisy and incomplete nature. We speculate this to be the reason behind lack of any prior noticeable work using textual description for improving instance segmentation. 

\begin{figure}[t]
	\centering
	\subfigure{\includegraphics[width=\linewidth]{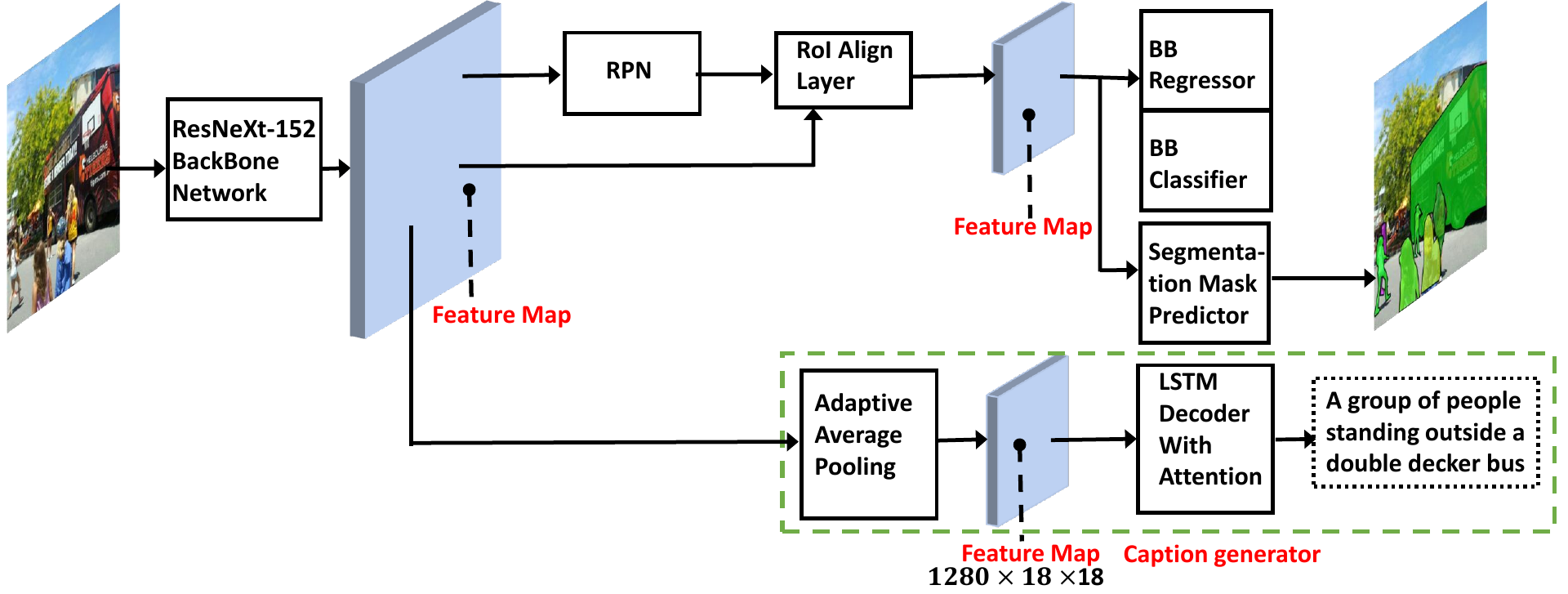}}
	\caption{Our framework for instance segmentation using Mask-RCNN and LSTM based caption generator to use textual descriptions of an image at test time.}
	% Note that BB refers to bounding box.}
	\label{fig:instance-seg-proposed-arch}
\end{figure}

\myparagraph{Our Implementation} 
Our MTL based architecture is shown is Figure~\ref{fig:instance-seg-proposed-arch}. We take Mask R-CNN~\cite{he2017mask}, a state-of-the-art instance segmentation approach, and combine it with the LSTM decoder from the captioning generator ``Show, Attend and Tell''(SAT)~\cite{xu2015show} within our framework. We use implementation of Mask R-CNN from \cite{mrcnn-code} (with ResNeXt-152 \cite{xie2017aggregated} backbone), and SAT from \cite{sat-code}. We initialize Mask R-CNN with pre-trained weights from \cite{mrcnn-weights}, and fine-tune with caption decoder (secondary task) using our MTL architecture. We have set early stopping iterations at 2, with  $\alpha=100$, and have used Adam optimizer with a learning rate of $5*10^{-6}$ at test time.

\setlength{\tabcolsep}{7pt}
\begin{table}
	\centering
	\begin{tabular}{llllll}
		\toprule[1.5pt]
		\textbf{Method} & \textbf{AP} & \textbf{AP\textsubscript{0.5}}&\textbf{AP\textsubscript{S}} & \textbf{AP\textsubscript{M}}& \textbf{AP\textsubscript{L}}\\
		\midrule
		MRCNN &  39.8 & 63.5  & 39.4 & 69.0  &  80.0 \\ 
		MR-MT &  39.9 & 63.6  & 39.2 & 69.2  &  79.8 \\ 
		\midrule
		%MM & ?? & ?? & ?? & ?? & ?? \\
		Ours-L2 &  \textbf{40.5} & \textbf{64.6} & \textbf{40.2} & \textbf{70.0}  & 81.2\\ 
		Ours-ES &  \textbf{40.5} &  \textbf{64.6} & 40.1 & \textbf{70.0}  & \textbf{82.1}    \\
		\bottomrule[1.5pt]
	\end{tabular}
\caption{Comparison of results for instance segmentation. AP\textsubscript{0.5} refers to AP at IoU of 0.5. AP\textsubscript{L}, AP\textsubscript{M},AP\textsubscript{S} represent AP\textsubscript{0.5} values for large, medium and small objects, respectively}
\label{table:instance_precision_comparison}
\end{table}

\myparagraph{Comparison}  Table~\ref{table:instance_precision_comparison} shows our comparison with various baselines. MRCNN is the original Mask RCNN instance segmentation model and MR-MT is the same model trained using MTL. 
%
%Uncomment the following if MM works
Similar to the case of semantic segmentation, we generate our own multi-modal network, referred to as MM. We gather the caption features generated by the state-of-the-art sentence embedding generator ``InferSent''~\cite{conneau2017supervised}, and append them to every position in height and width of an intermediate layer (i.e., feature map in Figure \ref{fig:instance-seg-proposed-arch}). The merged features are then passed through a $1\times1$ convolution to bring the feature size back to the original dimension. Rest of the network is same as the MRCNN. 
%During training, all the weights are kept fixed in the main pipeline, and only the weights for the $1 \times 1$ convolution layer are learnt.

%%%%%%%%%%%%%%%%%%%%%%%%
%other text for reference, do not uncomment even if MM works
%Similar to semantic segmentation experiments (section~\ref{sec:semantic_seg}), in case of MM caption features generated from state-of-the-art sentence embedding generator ``InferSent''~\cite{conneau2017supervised} are appended to the intermediate features (feature map in figure~\ref{fig:instance-seg-proposed-arch}) of the Mask R-CNN at every position in height and width. 
%We tried training the network in two settings (a) keeping the weights of the main pipeline fixed (b) fine-tuning them along with the new convolution layer weights. 
%training all the weights are fixed except for the $1\times 1$ convolution layer which we have introduced. 
%%%%%%%%%%%%%%%%%%%%%%%%

We compare with two versions of our approach with early stopping (ES) and L2 norm regularizers. Our attempts with pruning based approaches did not yield any gain in this case, since learning the label set to be pruned (using captions) turned out to be quite difficult. MM based approach also did not result in any performance improvement over the baselines; it is a difficult task for multi-modal approaches. We report our results on the validation set of MS-COCO \cite{lin2014microsoft}. We use AP (average precision) as our evaluation metric, and evaluate at AP\textsubscript{0.5} (IoU threshold of 0.5). AP\textsubscript{L}, AP\textsubscript{M},AP\textsubscript{S} represent AP\textsubscript{0.5} values for large, medium and small objects, respectively.

\myparagraph{Results on Small Objects}
The table~\ref{table:instance_object_comparison} presents the comparison for detecting various objects 
\setlength{\tabcolsep}{6pt}
\begin{table}
	\centering
	\begin{tabular}{lllllll}
		\toprule[1.5pt]
		\textbf{Type} & \multicolumn{3}{c}{\textbf{MR-MT}} & \multicolumn{3}{c}{\textbf{Ours-ES}} \\
		\midrule
		& \textbf{P}           & \textbf{R}         & \textbf{F1}          & \textbf{P}            & \textbf{R}           & \textbf{F1}           \\
		\midrule
		All           & 86.1  &  51.7 &  64.6   &   83.6 &  54.6 &  66.1       \\
		Small          & 80.6  & 29.8 &  43.5  &  77.9 & 31.9 & 45.3        \\
		Medium        & 86.6  & 61.3 &  71.8  & 83.9 &  64.5 &  72.9       \\
		Large         & 89.8 & 76.5 &  82.6 &  87.6 & 80.6 &  84.0   \\
		\bottomrule[1.5pt]   
	\end{tabular}
\caption{Comparison of baseline(MR-MT) and proposed(Ours-ES) approach at 0.5 IoU and 0.9 confidence threshold.  P: Precision, R: Recall, F1: F-measure }
\label{table:instance_object_comparison}
\end{table}
in terms of precision (P), recall (R) and f-measure (F1), at 0.5 IoU and 0.9 confidence threshold. 
We observe the maximum gain for small objects using our method, which makes sense since a large number of them remain undetected in the original model. Careful analysis revealed that ground truth itself has inconsistencies missing several smaller objects, undermining the actual gain obtained by our approach. We give the detailed analysis in the supplementary material.

\begin{figure}[t]
	\centering
	\subfigure{\includegraphics[width=0.99\linewidth]{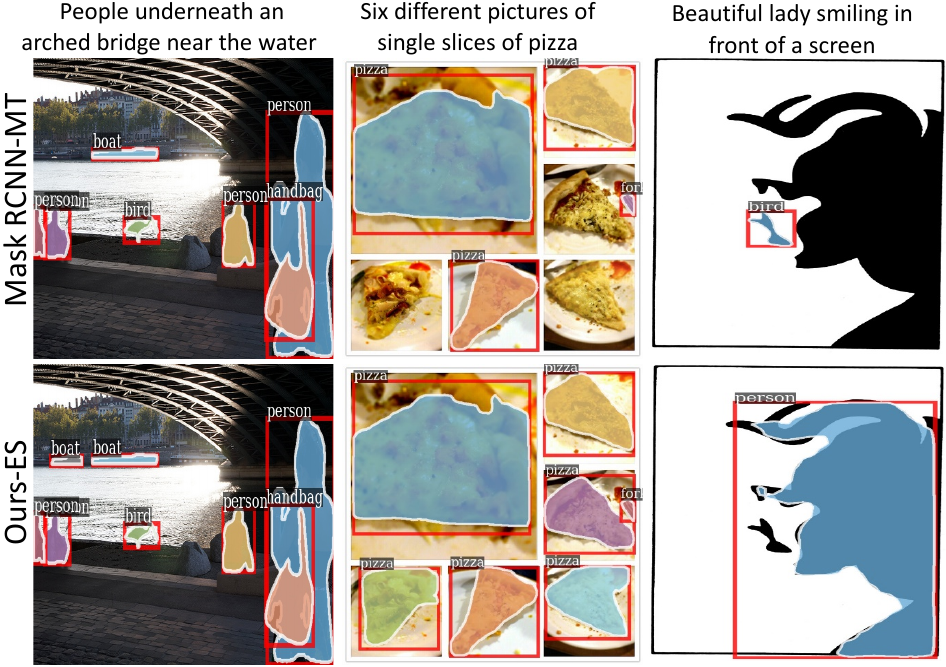}}
	\caption{Comparative results for instance segmentation given image description. Text at the top top of each column, shows the given image caption. Our approach is able to discover new objects (and their segmentation) which were missed earlier.}
	\label{fig:instance-seg-vis-results-main}
\end{figure}

\myparagraph{Visual Results}
Figure \ref{fig:instance-seg-vis-results-main} shows visual output from our framework. %We note that 
Our algorithm can detect newer objects, as well as detect additional objects of the same category. At times it is able to detect the objects which are not even mentioned in the caption, but are merely correlated with the caption.

%% file: sections/conclusion.tex
\section{Conclusion}\label{sec:conclusion}

We have presented a novel principled approach to perform evidence based conditional inference in DNNs. Our key idea is to model the evidence as auxiliary information in an MTL architecture and then modify the latent feature vector at prediction time such that the output of auxiliary task(s) matches the evidence. The approach is generic and can be used to improve the accuracy of existing DNN architectures for variety of tasks, using easily available auxiliary information at the test time. Experiments on three different vision applications demonstrate the efficacy of our proposed model over state-of-the-art. %We will publicly release all our code/models after paper acceptance.
%In future, we would like to experiment with additional applications such as for the video data.

%% file: sections/suppl.tex
\onecolumn
\title{\hspace{4cm}\fontsize{25}{30}\selectfont Supplementary Material}

\def\resultimgwidth{0.25\linewidth}
\def\resultimgheight{0.15\linewidth}
\setlength{\tabcolsep}{3pt}
\vspace{0.5cm}

\setcounter{section}{0}
\setcounter{figure}{0}
\setcounter{page}{1}

\section{Semantic Segmentation}

\subsection{Visual Results}

\begin{figure*}[h]
	\centering
	\subfigure{\includegraphics[width=\linewidth]{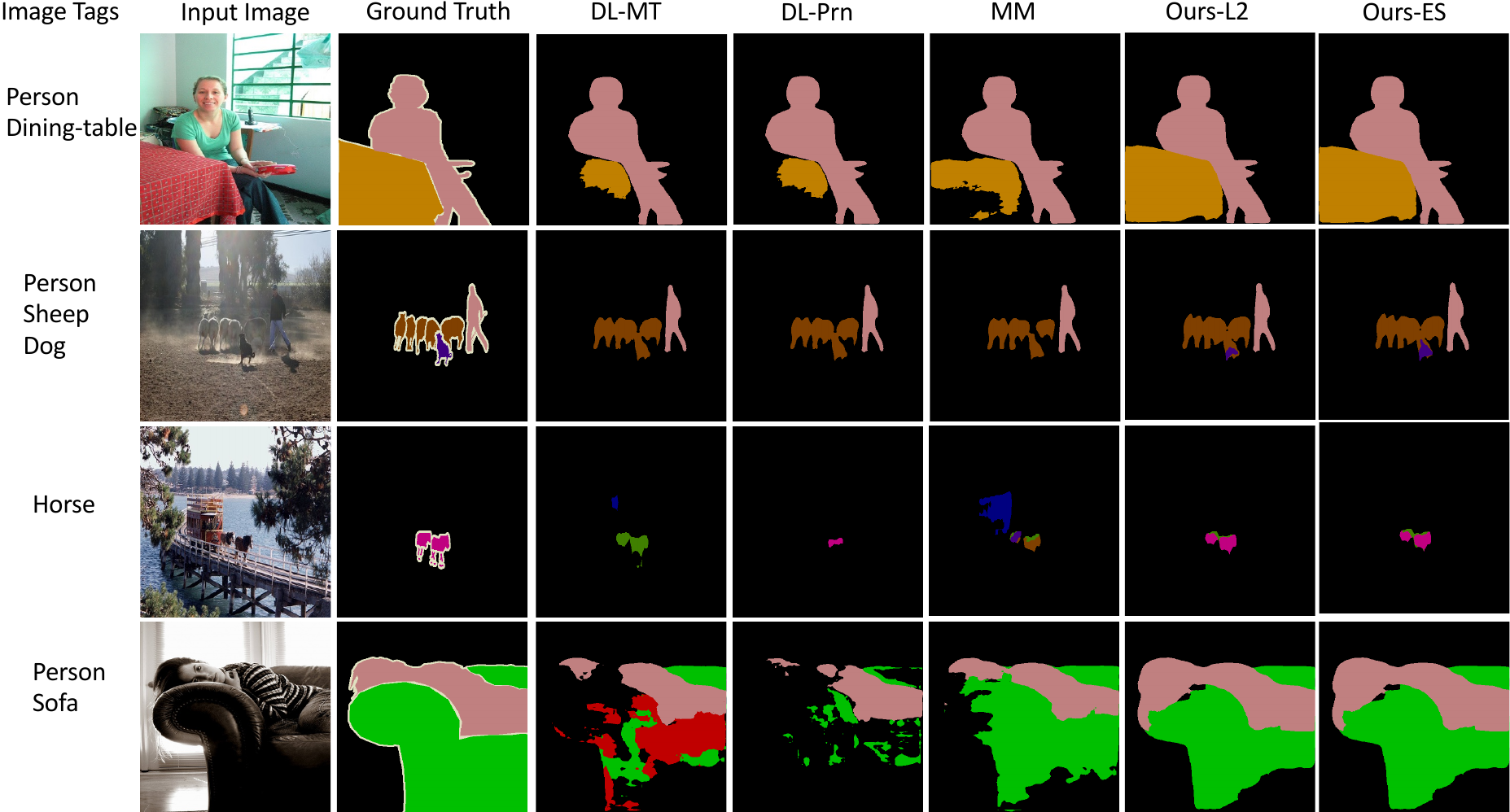}}
	\caption{Comparison of visual results on semantic segmentation problem using image level tags as the test time auxiliary information. 1st and 2nd column are input and ground truth respectively. 3rd column shows baseline model trained with MTL. 4th column shows results of a naive strategy to improve the results of 3rd column by pruning predicted labels which were not present in tags. 5th column shows results with our designed multi-modal network (MM). The last 2 columns show results of proposed approach in various configurations (see paper for details). The first row shows improvement in segmentation of `dining table'. Second row shows correction of `dog' label. Third row shows correction of the `horse' label and improvement in its segmentation. Fourth row shows significant improvements in segmentation of both `person' and `sofa'. }
	%Whereas MM (incorrectly) labels the entire lower region as `sofa', we (ours-ES2) are somewhat more cautious}
	% no space for this!:  sometimes even correcting the entire output, e.g., image with the tag rabbit was incorrectly segmented as a different object by the baseline.}
	%We can not only improve segmentation but also correct object labels discover objects correct labels which were incorrectly predictedobjects which }
	\label{fig:sem-seg-vis-output-supple}
\end{figure*}

Figure \ref{fig:sem-seg-vis-output-supple} presents some examples of the output of our framework for the problem of semantic segmentation with image level tags as auxiliary information. Here, we have taken some examples already presented in the text earlier (Figure 6 in the main paper) with additional comparisons, as well as presented results over some additional examples. Baseline methods as well as other competing approaches, including pruning based and multi-modal techniques are also compared.
%the alternate approaches and the baseline state-of-the-art techniques.

\subsection{Sensitivity Analysis}

\begin{figure}[h]
	\centering
	\subfigure{\includegraphics[height=0.3\linewidth]{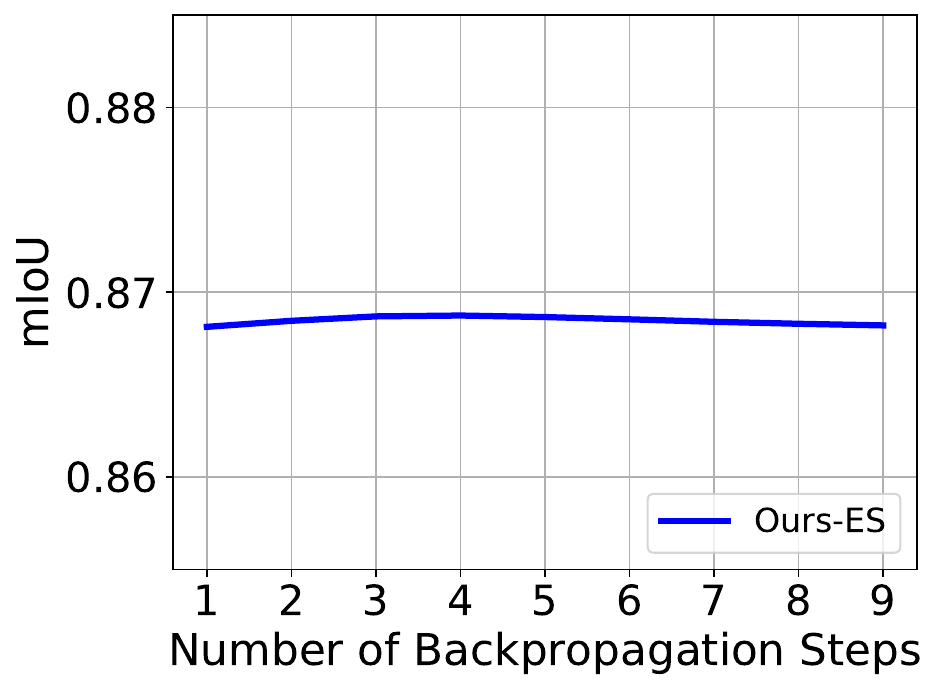}} \;
	\subfigure{\includegraphics[height=0.30\linewidth]{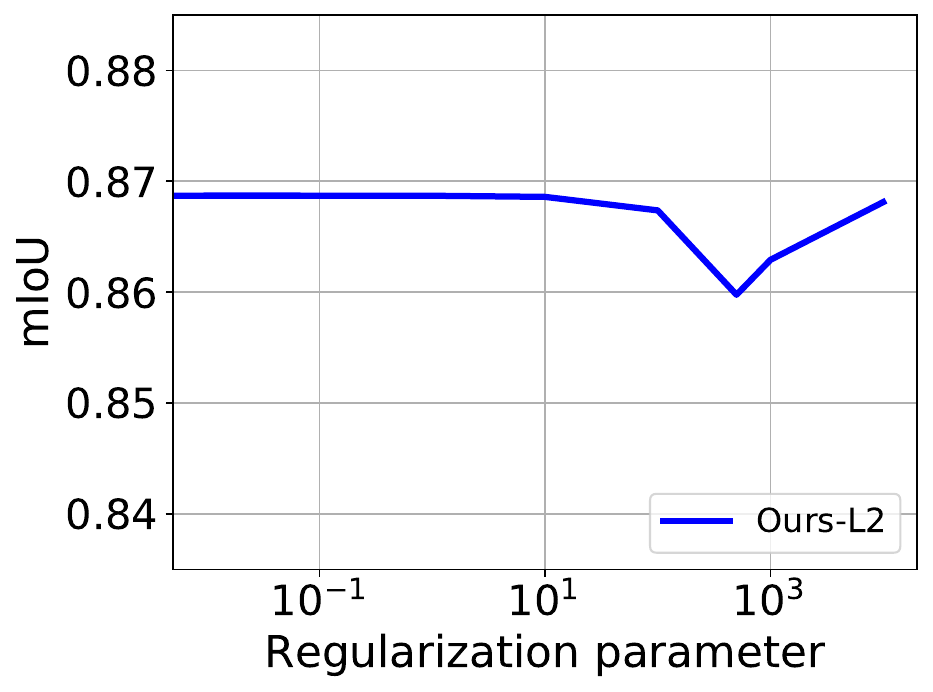}}		
	\caption{Sensitivity of results on number of back-propagation iterations (early stopping) and $\alpha$ parameter (L2-norm) in case of semantic segmentation using image level tags as auxiliary information. MS (multi-scaling) was kept on in this experiment.}
	\label{fig:sem-seg-sensitivity}	
\end{figure}

Figure \ref{fig:sem-seg-sensitivity} presents the sensitivity analysis with respect to the number of early stopping iterations (left) and the parameter controlling the weight of the L2 regularizer during prediction (right) in case of semantic segmentation using image level tags as auxiliary information. There is a large range of values where we get significant improvements over the baselines as well as other competing methods. These results were obtained with MS (multi-scaling) feature on.

\section{Instance Segmentation}

\subsection{Comparison with Pruning Based and Multi-Modal Approaches}
For the task of instance segmentation with captions as auxiliary information, we would like to make some additional remarks about our results presented in Table 4 of the main paper, and comparison with Pruning based and Multi-modal approaches.
%baselines. This is for the task of instance segmentation while using captions as auxiliary information. 

\myparagraph{Pruning Based} It is significantly harder to get pruning based methods to work in this case, since captions can often be noisy and do not always capture the precise information about which objects might (not) be present in the image. Hence, simply pruning based on objects which are not mentioned in the caption is likely not to give good results.
%They can often be noisy, and even contain object mentions which are not actually present in the image (if this line is not true, we can remove it). 
There is also an issue of matching the reference to an object in the image, since object labels are coming from a fixed set, whereas caption text is freely generated.
%(cite example from our visual results). 
 We tried a combination of various approaches, including (1) we generated a mapping of category labels to the words present in various captions, and then pruned based on this mapping (b) we tried expanding the set of words by including a list of commonly co-occurring words, and then pruning based on this expanded set. But none of these was able to improve the performance of the baseline, rather results became worse (and hence, are not reported in the text).

\myparagraph{Multi-modal} Designing a multi-modal (MM) architecture for the task of instance segmentation with captions (or even tags) is a significantly hard task, since instance segmentation pipeline has 3 parts to it (feature extraction, detection, segmentation), and it is not entirely clear where/how the caption based features should be merged with the instance segmentation pipeline. To the best of our knowledge, there is no existing neural architecture dealing with these two modes as input. Most natural was to merge the caption features in the beginning of the instance segmentation pipeline, while computing the first level of feature representation. The details of our multi-modal architecture are described in Section 4.3 (Comparison). We trained the multi-modal architecture for more than a week on a GPU, but the resulting (best) model could only achieve the accuracy which is at par with the MRCNN baseline. For comparison, each of our approaches' result (Ours-ES \& Ours-L2) was obtained by performing a training which lasted at most 4 days. In all cases (MM or Ours), we started with the pre-trained weights which came with the original MRCNN pipeline, since training the weights from scratch would be an overkill (it takes more than 3 weeks to train the MRCNN network). The multi-modal numbers are not reported in the table since they are in fact, almost identical to the MRCNN baseline.

In summary, we would like to stress that even designing a good enough competitor for our approach is a fairly difficult task in the above setting. There is no prior work which has done this in the neural setting taking multiple modes as input, and our own sincere attempts to do so did not yield any performance gains. Hence, improvement in results using our perturbation based method via back-propagation achieves additional significance, and novelty in this setting.

\subsection{Additional Visual Results}

\begin{figure*}[h]
	\centering
	\subfigure{\includegraphics[width=\linewidth]{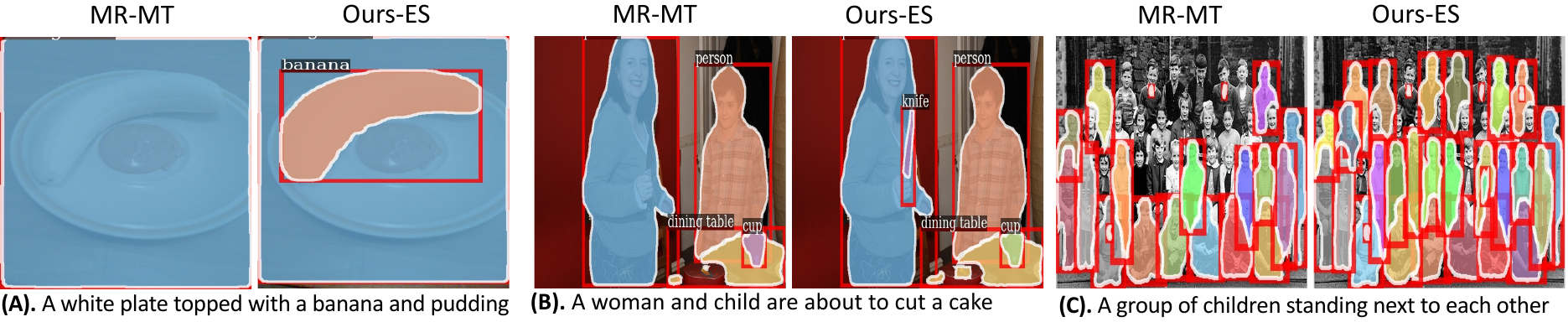}}
	\caption{Comparison of visual results for the baseline (MR-MT) and the proposed approach (Ours-ES) for the task of instance segmentation while using image captions as auxiliary information. Our approach is able to discover new objects (and their segmentation) which were missed earlier.}
	\label{fig:instance-seg-vis-results}
\end{figure*}

Figure \ref{fig:instance-seg-vis-results} presents some additional visual results for the problem of instance segmentation while using image captions as auxiliary information. We show a comparison between proposed (Ours-ES) and baseline (MR-MT) approaches.
%illustrates few more handpicked examples in addition to the examples presented in the paper in case of instance segmentation given captions of the image as the test time auxiliary information. It gives comparison of visual results between baseline (MR-MT) vs proposed (Ours-ES) approach. 
Our algorithm can detect newer objects (Figure~\ref{fig:instance-seg-vis-results} (A,B)), sometimes those not even mentioned in the caption (Figure~\ref{fig:instance-seg-vis-results} (B)), as well as detect additional objects of the same category (Figure~\ref{fig:instance-seg-vis-results} (C)).
%for the output from our framework and compares them with the alternate approaches and the baseline state of the art techniques.

\subsection{Inconsistencies in the Ground Truth Annotations}
We notice that in the case of instance segmentation, though our results are much better qualitatively, the same is not fully reflected in the quantitative comparison. A careful analysis of the results reveals that there are often inconsistencies in the ground truth annotations itself. For example, in many cases ground annotation misses objects which either very small, occluded or partly present in the image. In some cases, the ground truth segmentation has inconsistencies.

In many of these scenarios, our framework is able to predict the correct output, but since the ground truth annotation is erroneous, we are incorrectly penalized for detecting true objects or for detecting the correct segmentation, resulting in apparent loss in our accuracy numbers. In the figures below, we highlight some examples to support our claim. Systematically fixing the ground truth is a direction for future work.
\vspace{1em}
\begin{figure*}[h]
	\renewcommand{\tabcolsep}{0.04cm}
	\centering
	\begin{tabular}{cccc}
		\includegraphics[width=\resultimgwidth,height=\resultimgheight]{./figures/supplementary/5503.jpg}
		&
		\includegraphics[width=\resultimgwidth,height=\resultimgheight]{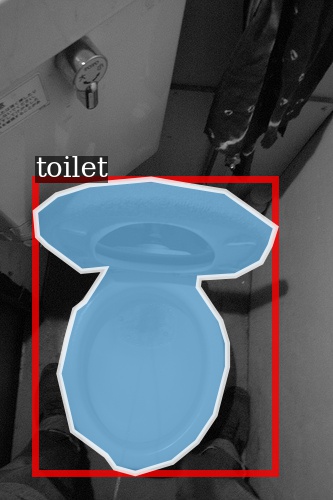}
		&
		\includegraphics[width=\resultimgwidth,height=\resultimgheight]{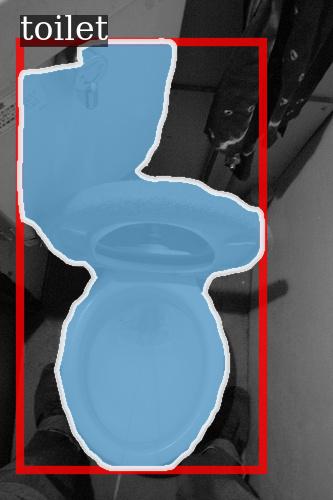}
		&              \includegraphics[width=\resultimgwidth,height=\resultimgheight]{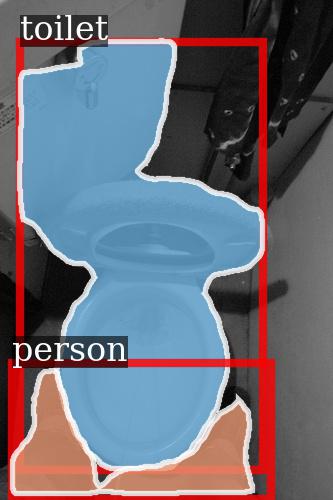}
		\\
		
		\multirow{2}{*}{\multilineC{Input Image}} &
		\multirow{2}{*}{\multilineC{Ground Truth}} &
		\multirow{2}{*}{\multilineC{MR-MT Output}}
		&\multirow{2}{*}{\multilineC{Ours-ES Output \\ (Our Approach)}}

	\end{tabular}
	\caption{Input caption: ``a person standing over a toilet using the restroom''. \\
	Ours-ES is able to detect the person standing in the toilet. This is in contrast to MR-MT which can only detect the toilet. Ground truth incorrectly misses the person.}
	
%A cat looking at another cat on the television''. \\ In the ground truth both the cats are labeled as one, but Mask-Aux-L2 is able to predicts two cats while Mask-MTL predicts only one cat.}
	\label{fig:toilet}
\end{figure*}
\vspace{1em}
\begin{figure*}[h]
	\renewcommand{\tabcolsep}{0.04cm}
	\centering
	\begin{tabular}{cccc}
		\includegraphics[width=\resultimgwidth,height=\resultimgheight]{./figures/supplementary/532761.jpg}
		&
		\includegraphics[width=\resultimgwidth,height=\resultimgheight]{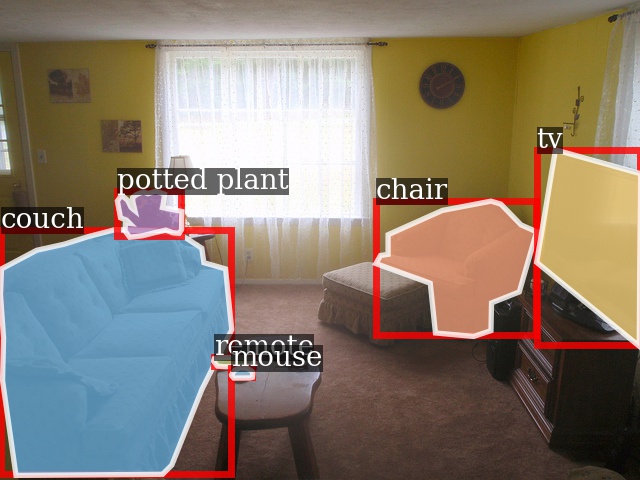}
		&
		\includegraphics[width=\resultimgwidth,height=\resultimgheight]{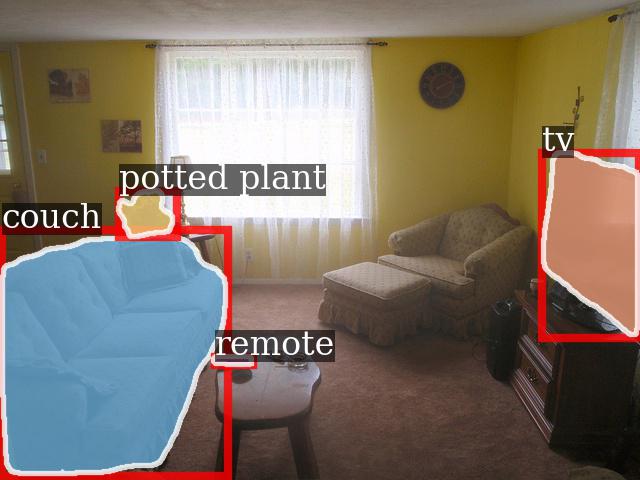}
		&              \includegraphics[width=\resultimgwidth,height=\resultimgheight]{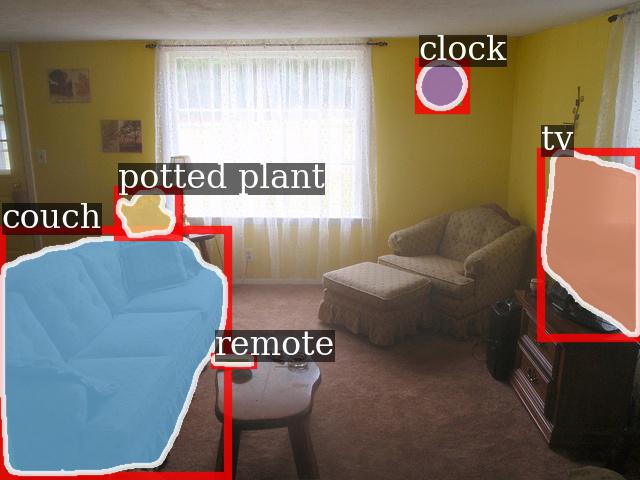}
		\\
		
		\multirow{2}{*}{\multilineC{Input Image}} &
		\multirow{2}{*}{\multilineC{Ground Truth}} &
		\multirow{2}{*}{\multilineC{MR-MT Output}}
		&\multirow{2}{*}{\multilineC{Ours-ES Output \\ (Our Approach)}}

	\end{tabular}
	\caption{Input caption: ``a living room with a couch, coffee table, tv on a stand and a chair with ottoman''. \\ Ours-ES is able to detect clock as compared to MR-MT. This is not annotated in the ground truth.} %so treated as a false positive.}
	\label{fig:clock}
\end{figure*}
\vspace{1em}
\begin{figure*}[h]
	\renewcommand{\tabcolsep}{0.04cm}
	\centering
	\begin{tabular}{cccc}
		\includegraphics[width=\resultimgwidth,height=\resultimgheight]{./figures/supplementary/557884.jpg}
		&
		\includegraphics[width=\resultimgwidth,height=\resultimgheight]{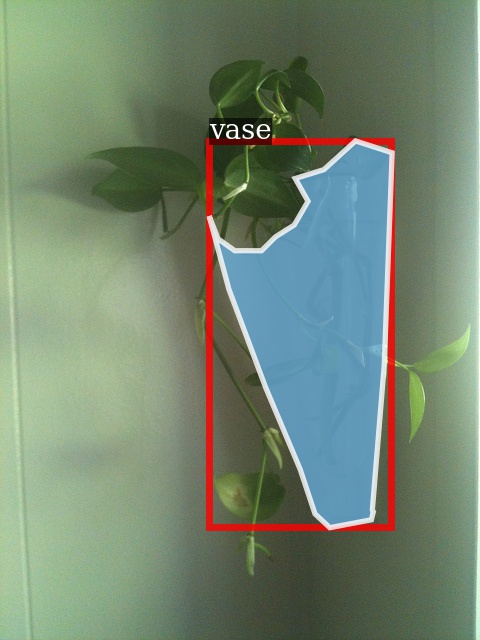}
		&
		\includegraphics[width=\resultimgwidth,height=\resultimgheight]{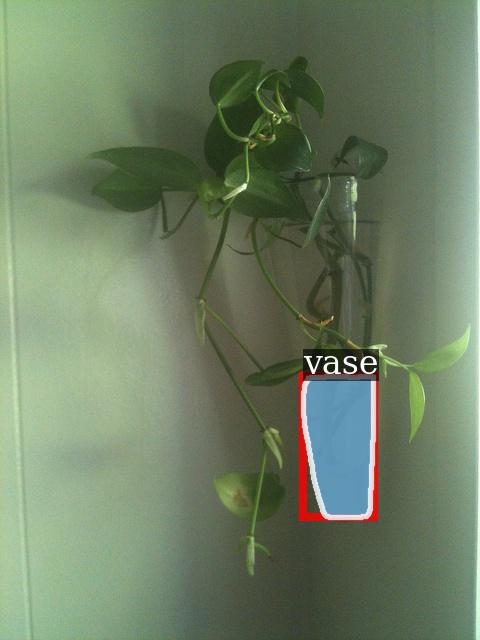}
		&              \includegraphics[width=\resultimgwidth,height=\resultimgheight]{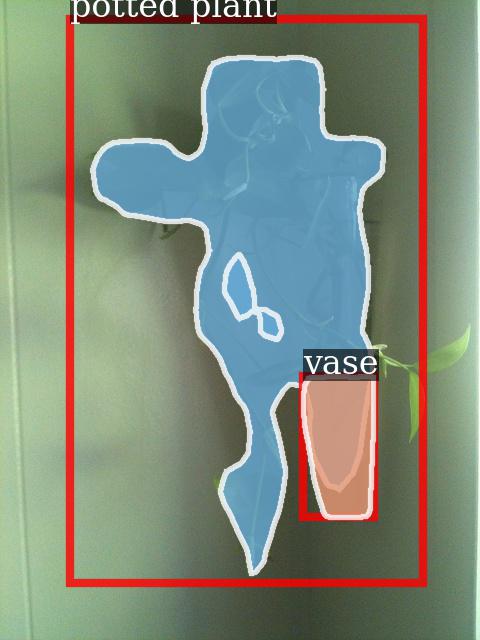}
		\\
		
		\multirow{2}{*}{\multilineC{Input Image}} &
		\multirow{2}{*}{\multilineC{Ground Truth}} &
		\multirow{2}{*}{\multilineC{MR-MT Output}}
		&\multirow{2}{*}{\multilineC{Ours-ES Output \\ (Our Approach)}}

	\end{tabular}
	\caption{Input caption: ``the large glass plant vase is installed into the wall''. \\ Ours-ES is able to detect potted plant as compared to MR-MT. This is not annotated in the ground truth.} %so treated as a false positive.}
	\label{fig:potted_plat}
\end{figure*}

\vspace{1em}

\begin{figure*}[h]
	\renewcommand{\tabcolsep}{0.04cm}
	\centering
	\begin{tabular}{cccc}
		\includegraphics[width=\resultimgwidth,height=\resultimgheight]{./figures/supplementary/547854.jpg}
		&
		\includegraphics[width=\resultimgwidth,height=\resultimgheight]{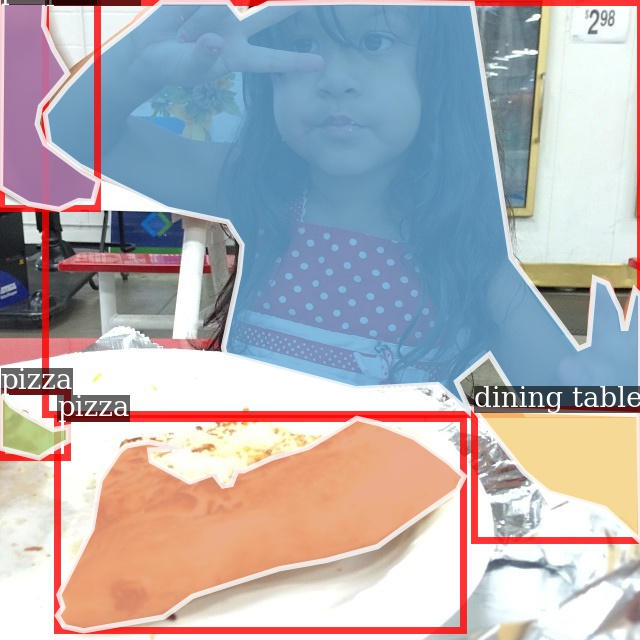}
		&
		\includegraphics[width=\resultimgwidth,height=\resultimgheight]{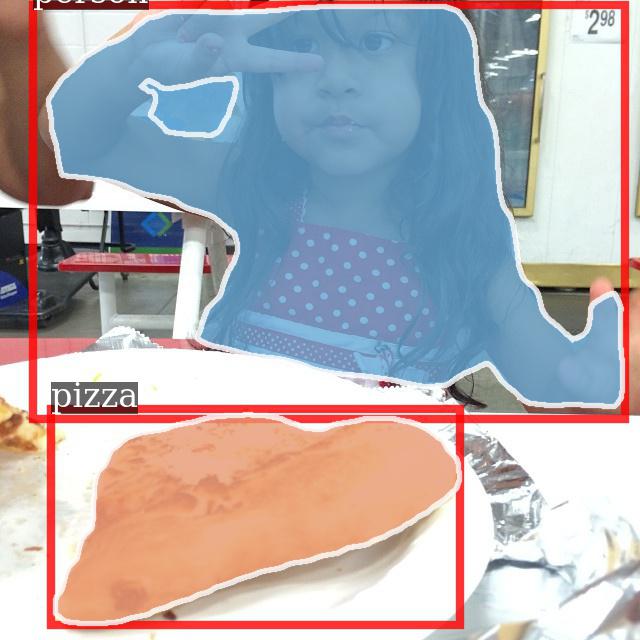}
		&              \includegraphics[width=\resultimgwidth,height=\resultimgheight]{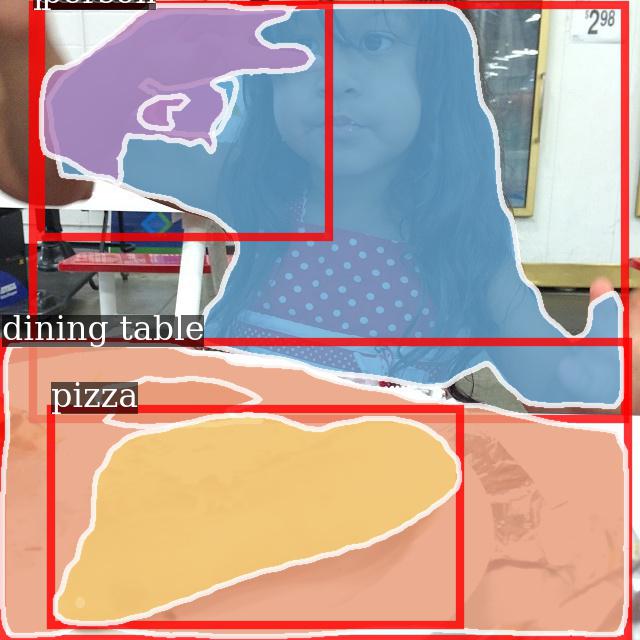}
		\\
		
		\multirow{2}{*}{\multilineC{Input Image}} &
		\multirow{2}{*}{\multilineC{Ground Truth}} &
		\multirow{2}{*}{\multilineC{MR-MT Output}}
		&\multirow{2}{*}{\multilineC{Ours-ES Output \\ (Our Approach)}}

	\end{tabular}
	\caption{Input caption: ``the little girl is busy eating her pizza at the table''. \\ Ours-ES is able to detect dining table as compared to MR-MT. The dining table is labeled in the ground truth with incomplete segmentation.} %so treated as a false positive.}
	\label{fig:dining_table}
\end{figure*}

\vspace{1em}

\begin{figure*}[h]
	\renewcommand{\tabcolsep}{0.04cm}
	\centering
	\begin{tabular}{cccc}
		\includegraphics[width=\resultimgwidth,height=\resultimgheight]{./figures/supplementary/537802.jpg}
		&
		\includegraphics[width=\resultimgwidth,height=\resultimgheight]{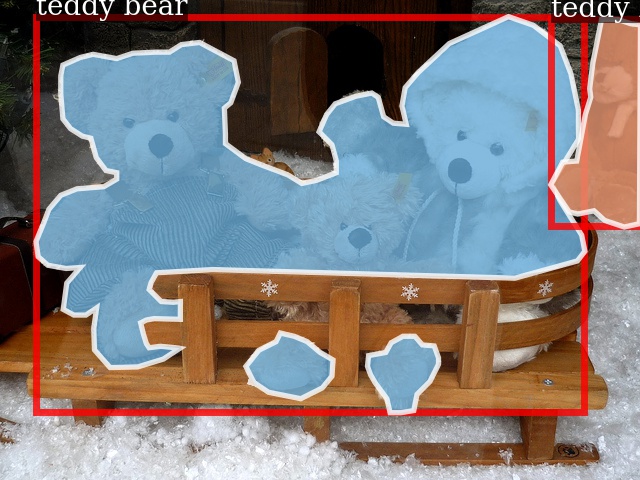}
		&
		\includegraphics[width=\resultimgwidth,height=\resultimgheight]{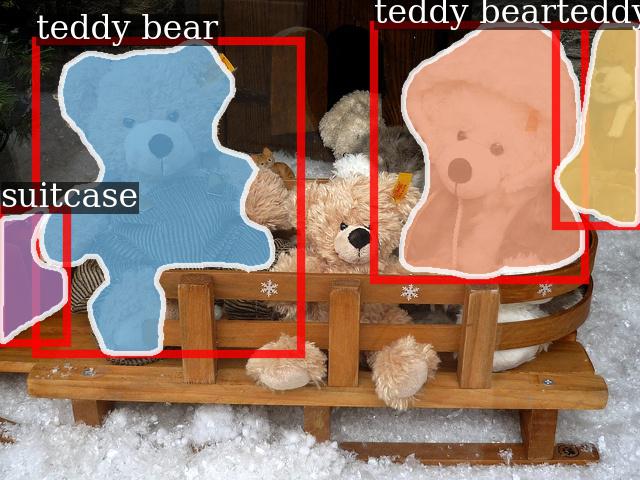}
		&              \includegraphics[width=\resultimgwidth,height=\resultimgheight]{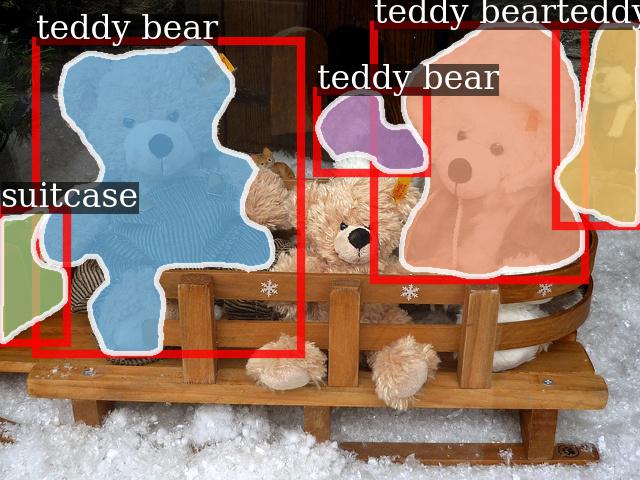}
		\\
		
		\multirow{2}{*}{\multilineC{Input Image}} &
		\multirow{2}{*}{\multilineC{Ground Truth}} &
		\multirow{2}{*}{\multilineC{MR-MT Output}}
		&\multirow{2}{*}{\multilineC{Ours-ES Output \\ (Our Approach)}}

	\end{tabular}
	\caption{Input caption: ``three teddy bears sit in a sled in fake snow''. \\ Ours-ES is able to detect 4 teddy bears as compared to MR-MT which detects 3 teddy bears. All 3 teddy bears in the front portion of image are marked as a single teddy bears in the ground truth.} % and so treated as false positives.}
	\label{fig:teddy}
\end{figure*}

\vspace{1em}

\begin{figure*}[h]
	\renewcommand{\tabcolsep}{0.04cm}
	\centering
	\begin{tabular}{cccc}
		\includegraphics[width=\resultimgwidth,height=\resultimgheight]{./figures/supplementary/535253.jpg}
		&
		\includegraphics[width=\resultimgwidth,height=\resultimgheight]{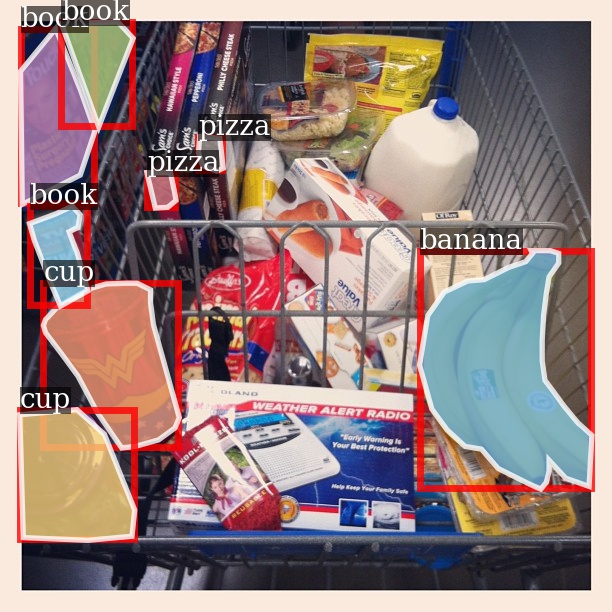}
		&
		\includegraphics[width=\resultimgwidth,height=\resultimgheight]{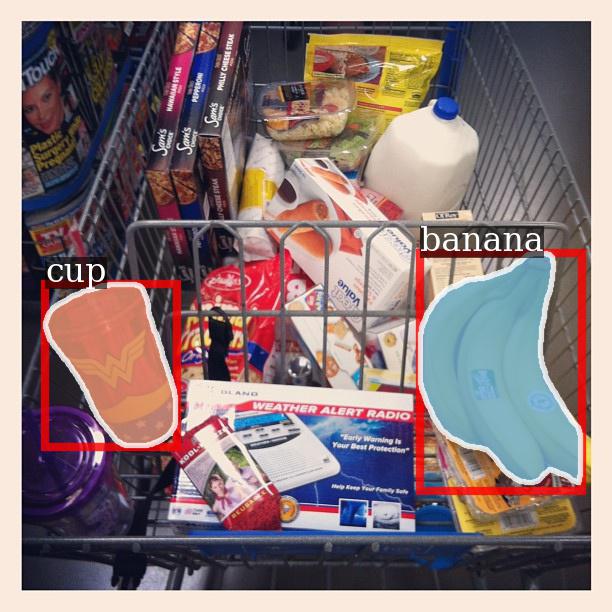}
		&              \includegraphics[width=\resultimgwidth,height=\resultimgheight]{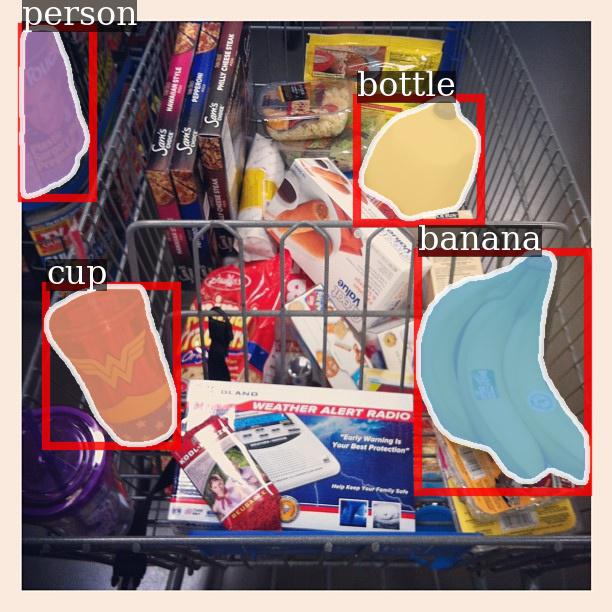}
		\\
		
		\multirow{2}{*}{\multilineC{Input Image}} &
		\multirow{2}{*}{\multilineC{Ground Truth}} &
		\multirow{2}{*}{\multilineC{MR-MT Output}}
		&\multirow{2}{*}{\multilineC{Ours-ES Output \\ (Our Approach)}}

	\end{tabular}
	\caption{Input caption: ``a shopping cart full of food that includes bananas and milk''. \\ Ours-ES is able to detect bottle of milk as compared to MR-MT. This is not annotated in the ground truth.} % and so treated as false positives.}
	\label{fig:bottle}
\end{figure*}

\vspace{1em}

\begin{figure*}[h]
	\renewcommand{\tabcolsep}{0.04cm}
	\centering
	\begin{tabular}{cccc}
		\includegraphics[width=\resultimgwidth,height=\resultimgheight]{./figures/supplementary/6012.jpg}
		&
		\includegraphics[width=\resultimgwidth,height=\resultimgheight]{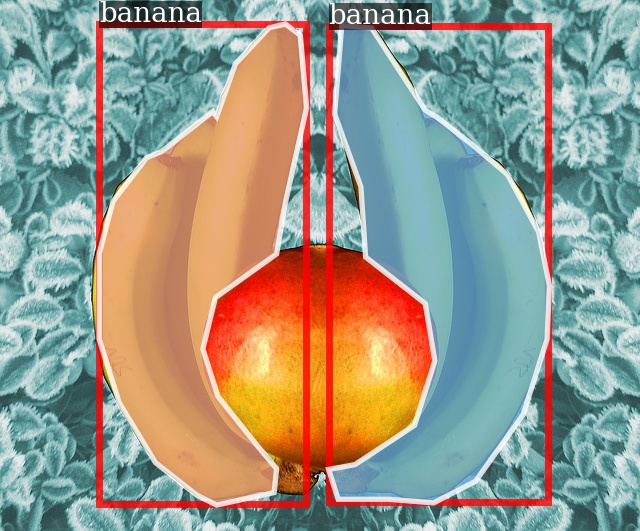}
		&
		\includegraphics[width=\resultimgwidth,height=\resultimgheight]{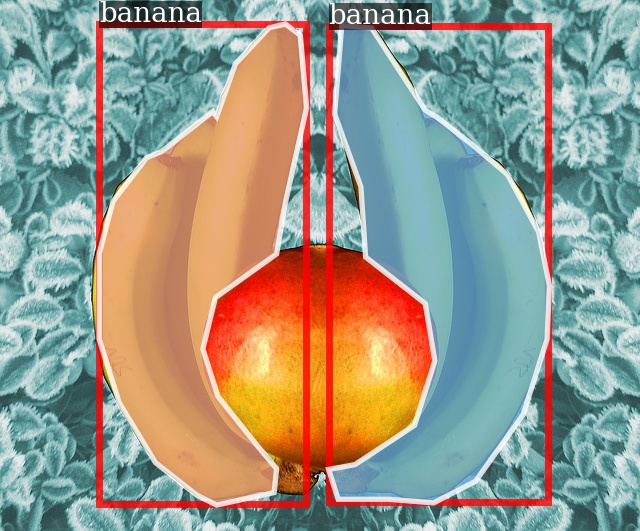}
		&              \includegraphics[width=\resultimgwidth,height=\resultimgheight]{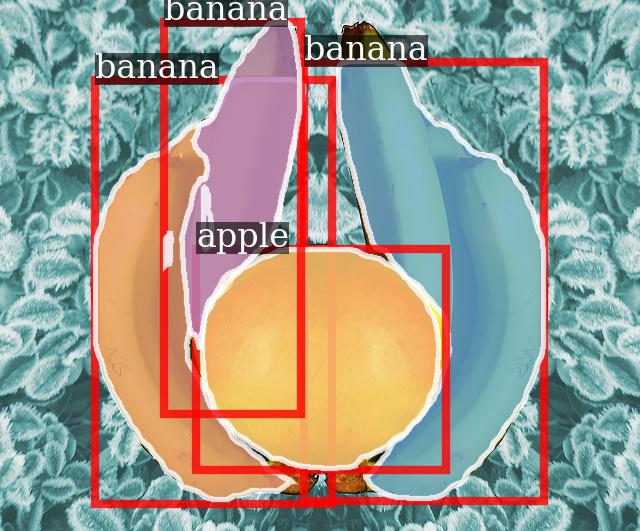}
		\\
		
		\multirow{2}{*}{\multilineC{Input Image}} &
		\multirow{2}{*}{\multilineC{Ground Truth}} &
		\multirow{2}{*}{\multilineC{MR-MT Output}}
		&\multirow{2}{*}{\multilineC{Ours-ES Output \\ (Our Approach)}}

	\end{tabular}
		\caption{Input caption: ``a yellow and red apple and some bananas''. \\ Ours-ES is able to separte the bananas as compared to MR-MT. These are marked as a single banana in the ground truth.} % and so treated as false positives.}
	\label{fig:bananas}
\end{figure*}
\vspace{1em}

\begin{figure*}[h]
	\renewcommand{\tabcolsep}{0.04cm}
	\centering
	\begin{tabular}{cccc}
		\includegraphics[width=\resultimgwidth,height=\resultimgheight]{./figures/supplementary/6614.jpg}
		&
		\includegraphics[width=\resultimgwidth,height=\resultimgheight]{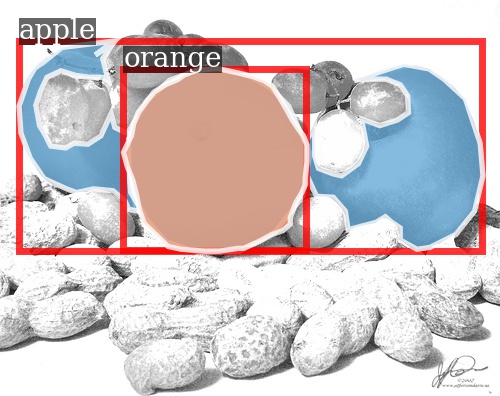}
		&
		\includegraphics[width=\resultimgwidth,height=\resultimgheight]{./figures/supplementary/6614_mt.jpg}
		&              \includegraphics[width=\resultimgwidth,height=\resultimgheight]{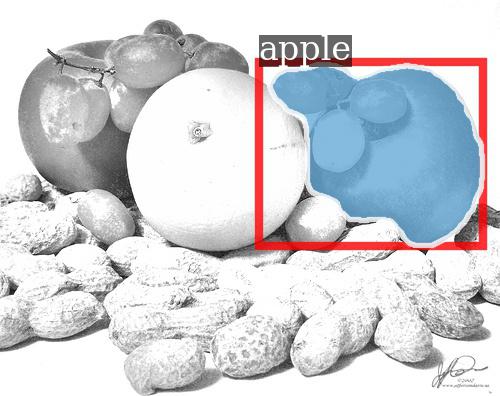}
		\\
		
		\multirow{2}{*}{\multilineC{Input Image}} &
		\multirow{2}{*}{\multilineC{Ground Truth}} &
		\multirow{2}{*}{\multilineC{MR-MT Output}}
		&\multirow{2}{*}{\multilineC{Ours-ES Output \\ (Our Approach)}}

	\end{tabular}
	\caption{Input caption: ``two apples, an orange, some grapes and peanuts''. \\ 
	Ours-ES is able to detect the an apple in the image as compared to MR-MT, but both the apples are marked as single apple in the ground truth.}
	%Mask-Aux-L2 is able to detect a piece of pizza as compared to Mask-MTL, which is not annotated in the ground truth.
	\label{fig:apple}
\end{figure*}

\vspace{1em}

\begin{figure*}[h]
	\renewcommand{\tabcolsep}{0.04cm}
	\centering
	\begin{tabular}{cccc}
		\includegraphics[width=\resultimgwidth,height=\resultimgheight]{./figures/supplementary/32811.jpg}
		&
		\includegraphics[width=\resultimgwidth,height=\resultimgheight]{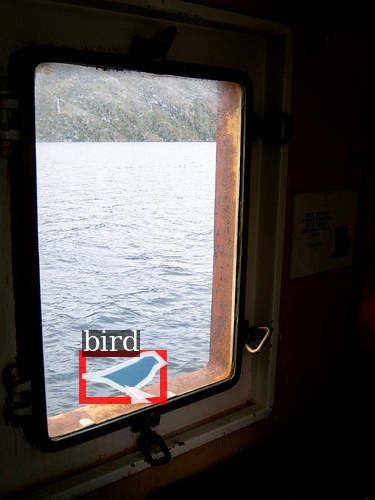}
		&
		\includegraphics[width=\resultimgwidth,height=\resultimgheight]{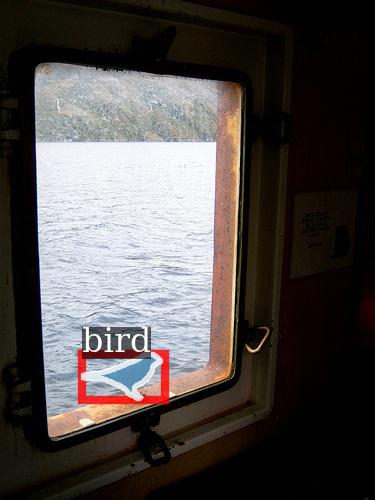}
		&              \includegraphics[width=\resultimgwidth,height=\resultimgheight]{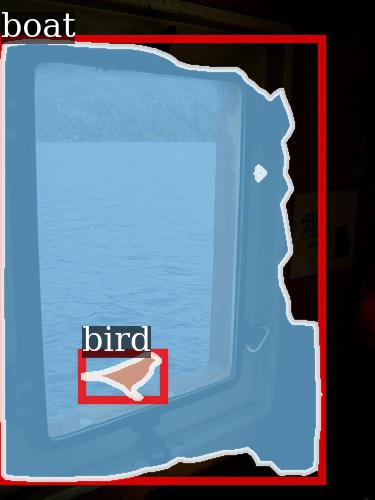}
		\\
		
		\multirow{2}{*}{\multilineC{Input Image}} &
		\multirow{2}{*}{\multilineC{Ground Truth}} &
		\multirow{2}{*}{\multilineC{MR-MT Output}}
		&\multirow{2}{*}{\multilineC{Ours-ES Output \\ (Our Approach) }}

	\end{tabular}
	\caption{Input caption: `a bird resting outside of a boat window''. \\Ours-ES is able to detect both bird and boat in the image as compared to MR-MT as mentined in the caption also. Ground truth has no annotaion for boat.}
	\label{fig:boat}
\end{figure*}

\vspace{1em}

\begin{figure}[H]
	\renewcommand{\tabcolsep}{0.04cm}
	\centering
	\begin{tabular}{cccc}
		\includegraphics[width=\resultimgwidth,height=\resultimgheight]{./figures/supplementary/136772.jpg}
		&
		\includegraphics[width=\resultimgwidth,height=\resultimgheight]{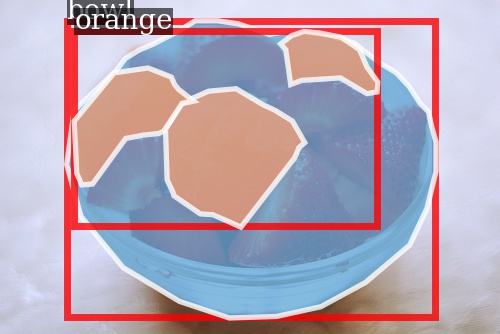}
		&
		\includegraphics[width=\resultimgwidth,height=\resultimgheight]{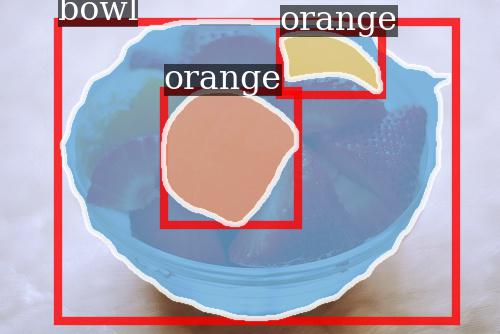}
		&              \includegraphics[width=\resultimgwidth,height=\resultimgheight]{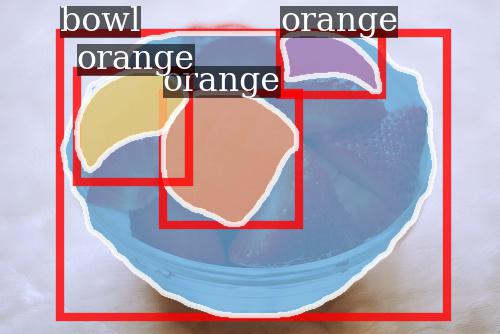}
		\\
		
		\multirow{2}{*}{\multilineC{Input Image}} &
		\multirow{2}{*}{\multilineC{Ground Truth}} &
		\multirow{2}{*}{\multilineC{MR-MT Output}}
		&\multirow{2}{*}{\multilineC{Ours-ES Output \\ (Our Approach)}}

	\end{tabular}
	\caption{Input caption: ``a lot of strawberries and oranges sitting in a bowl''. \\ Ours-ES is able to correctly detect an extra orange as compared to MR-MT. These all oranges are marked as a single orange in the ground truth.}
	\label{fig:oranges}
\end{figure}

\vspace{1em}

\begin{figure}[H]
	\renewcommand{\tabcolsep}{0.04cm}
	\centering
	\begin{tabular}{cccc}
		\includegraphics[width=\resultimgwidth,height=\resultimgheight]{./figures/supplementary/108026.jpg}
		&
		\includegraphics[width=\resultimgwidth,height=\resultimgheight]{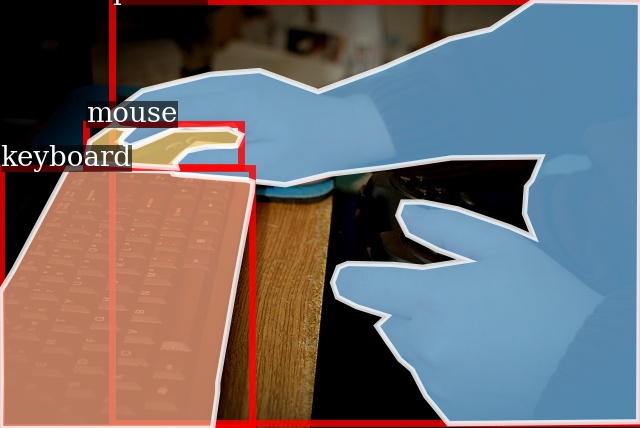}
		&
		\includegraphics[width=\resultimgwidth,height=\resultimgheight]{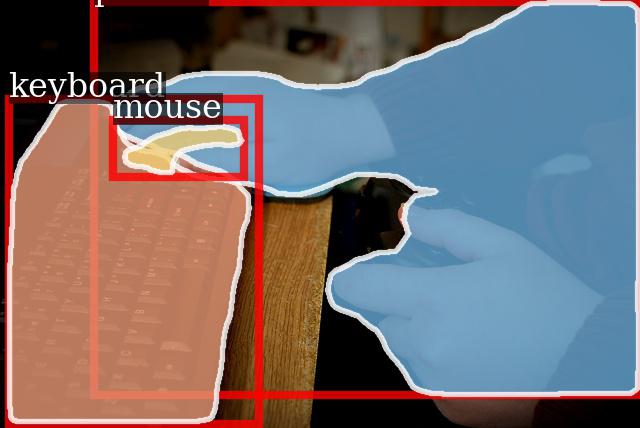}
		&              \includegraphics[width=\resultimgwidth,height=\resultimgheight]{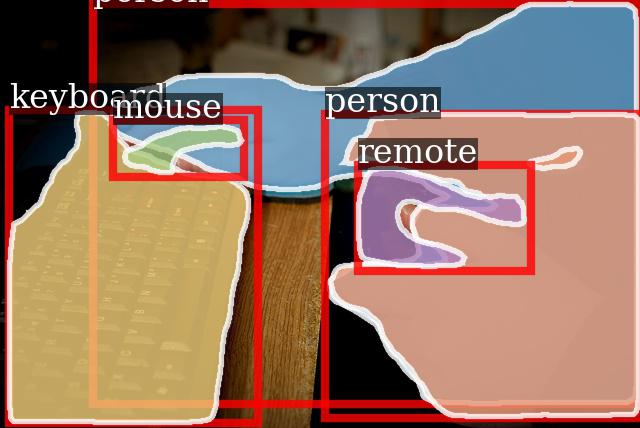}
		\\
		
		\multirow{2}{*}{\multilineC{Input Image}} &
		\multirow{2}{*}{\multilineC{Ground Truth}} &
		\multirow{2}{*}{\multilineC{MR-MT Output}}
		&\multirow{2}{*}{\multilineC{Ours-ES Output \\ (Our Approach)}}

	\end{tabular}
	\caption{Input caption: `a man holds a controller and keyboard with his hands''. \\ Ours-ES is able to detect the remote which is missed by MR-MT. The ground truth annotation misses the remote.}
	\label{fig:remote}
\end{figure}

\vspace{1em}

\section{Architecture For Using Multiple Evidence Simultaneously}

\begin{figure}[h]
 	\centering
 	\subfigure{\includegraphics[width=0.9\linewidth, height=0.5\linewidth]{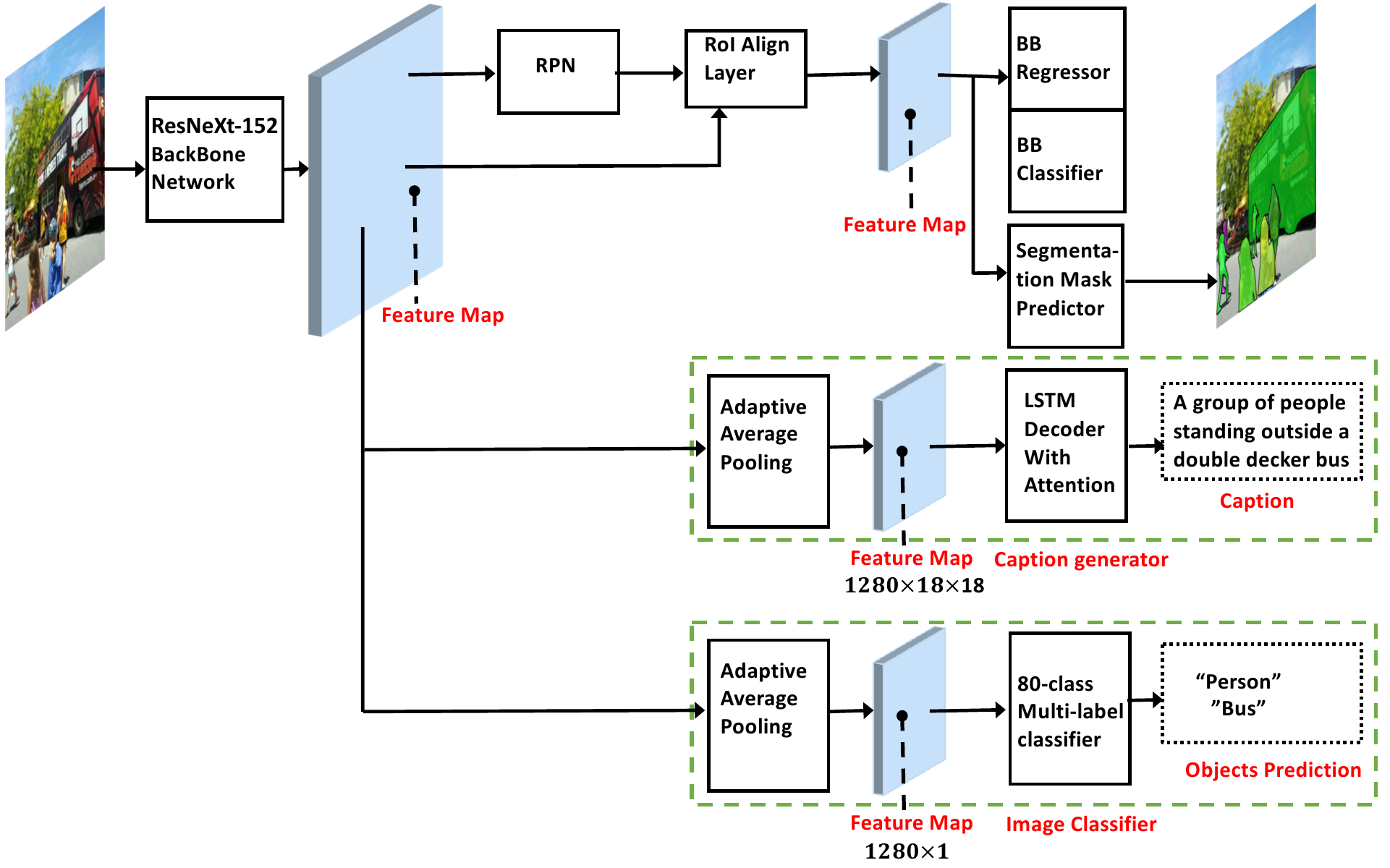}}
 	\caption{Our framework, for instance segmentation with multiple types of auxiliary information simultaneously. }
 	% Note that BB refers to bounding box.}
 	\label{fig:instance-seg-captions-tags-proposed-arch}
 \end{figure}

We have conducted experiments using multiple types of auxiliary information simultaneously in our instance segmentation setup. In addition to using captions as auxiliary information, we added image tags as the second source of auxiliary information. Figure~\ref{fig:instance-seg-captions-tags-proposed-arch} describes our proposed MTL architecture in detail.

\newpage

